\documentclass[runningheads]{llncs}
\usepackage{graphicx}

\usepackage{tikz}
\usepackage{comment}
\usepackage{amsmath,amssymb} 
\usepackage{color}
\usepackage{color}
\usepackage{amsmath}
\usepackage{amssymb}
\usepackage{multirow, varwidth}
\usepackage{epsfig}
\usepackage{verbatim}
\makeatletter
\newif\if@restonecol
\makeatother

\usepackage{xr}
\usepackage[amsmath,thmmarks]{ntheorem}
\usepackage{flushend}

\newcommand{\argmax}{\operatornamewithlimits{argmax}}

\DeclareRobustCommand\onedot{\futurelet\@let@token\@onedot}
\def\onedot{.} 
\def\eg{\emph{e.g}\onedot, }

 \def\vs{\emph{vs}\onedot}
 
\def\etal{\emph{et al}\onedot}

\usepackage{color}
\usepackage{graphicx}
\usepackage{amsmath}
\usepackage{amssymb}
\usepackage{booktabs}
\usepackage{kotex}
\usepackage{algorithm}
\usepackage[noend]{algorithmic}
\usepackage{mathtools}
\usepackage{amstext}
\usepackage{bbm}
\usepackage{tablefootnote}
\usepackage{makecell}
\usepackage{hyperref}
\hypersetup{final}
\usepackage{multicol}
\usepackage[accsupp]{axessibility}  
\usepackage{subcaption}

\usepackage{color} 
\usepackage[capitalise]{cleveref}
\usepackage{tabu}
\usepackage{wrapfig}

\makeatletter
\newcommand{\printfnsymbol}[1]{\textsuperscript{\@fnsymbol{#1}}}
\makeatother

\begin{document}
\sloppy
\pagestyle{headings}
\mainmatter
\def\ECCVSubNumber{5458}  

\title{Object Discovery via Contrastive Learning for Weakly Supervised Object Detection} 


\titlerunning{Object Discovery via Contrastive Learning for WSOD}
%
\author{Jinhwan Seo\inst{1}
\and
Wonho Bae\inst{2}
\and
Danica J.\ Sutherland\inst{2,3}
\and\\
Junhyug Noh\inst{4}\thanks{Corresponding authors: J. Noh (noh1@llnl.gov) and D. Kim (dkim@postech.ac.kr)}
\and
Daijin Kim\inst{1}\printfnsymbol{1}
} 
\authorrunning{J. Seo et al.}
%
\institute{Pohang University of Science and Technology\and
University of British Columbia\and
Alberta Machine Intelligence Institute\and
Lawrence Livermore National Laboratory
\\
\email{tohoaa@gmail.com, whbae@cs.ubc.ca, dsuth@cs.ubc.ca, \\noh1@llnl.gov, dkim@postech.ac.kr}\\
}
\maketitle

\begin{abstract}
Weakly Supervised Object Detection (WSOD) is a task that detects objects in an image using a model trained only on image-level annotations.
Current state-of-the-art models benefit from self-supervised instance-level supervision, but since weak supervision does not include count or location information, the most common ``argmax'' labeling method often ignores many instances of objects.
To alleviate this issue, we propose a novel multiple instance labeling method called \textit{object discovery}.
We further introduce a new contrastive loss under weak supervision where no instance-level information is available for sampling, called \textit{weakly supervised contrastive loss} (WSCL).
WSCL aims to construct a credible similarity threshold for object discovery by leveraging consistent features for embedding vectors in the same class.
As a result, we achieve new state-of-the-art results on MS-COCO 2014 and 2017 as well as PASCAL VOC 2012, and competitive results on PASCAL VOC 2007. The code is available at \url{https://github.com/jinhseo/OD-WSCL}.

\keywords{Weakly Supervised Object Detection (WSOD)}
\end{abstract}

\section{Introduction}
\label{sec:intro}
Object detection~\cite{ren2015faster,long2015fully,redmon2016you,liu2016ssd} has seen huge improvements since the introduction of deep neural networks and large-scale datasets~\cite{Everingham15,lin2014microsoft,deng2009imagenet}.
It is, however, very expensive and time-consuming to annotate large datasets with fine-grained object bounding boxes.
Recent work has thus attempted to use more cost-efficient annotations in an approach called Weakly Supervised Object Detection (WSOD),
such as image, point, or ``scribble'' labels.

\begin{figure}[t!]
\centering
\includegraphics[trim=0cm 0cm 0cm 0cm,clip,width=0.95\textwidth]{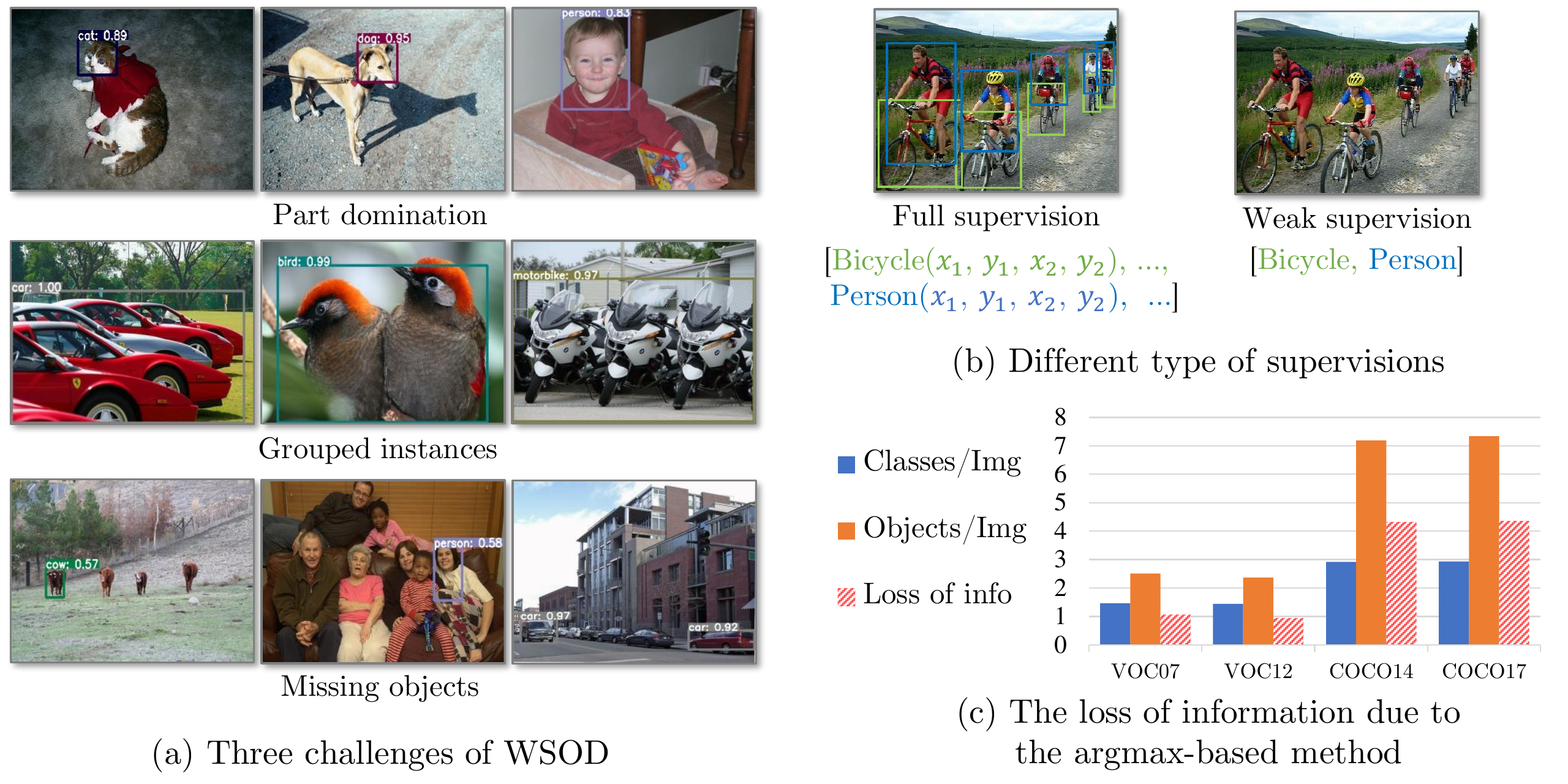}
\caption{(a) Three challenges of WSOD: part domination, grouped instances, and missing objects. (b) Unlike full supervision, there is no location and count information in weak supervision. (c) A large number of target objects are ignored in PASCAL VOC~\cite{Everingham15} and MS-COCO~\cite{lin2014microsoft} due to the argmax-based method.}
\label{fig:figure1}
\end{figure}

Although WSOD methods can be trained with much less annotation effort, however, the resulting models still perform far below their fully-supervised counterparts.
We identify three categories of reasons for this deterioration, summarized in \cref{fig:figure1}(a).
First, \emph{part domination} is when WSOD models focus only on the discriminative part of an object,
perhaps caused by the fundamentally ill-posed nature of framing WSOD as a Multiple Instance Learning (MIL) problem \cite{dietterich1997solving} prone to local minima,
as done by much previous work \cite{Bilen_2016_CVPR,Tang_2017_CVPR,Wan_2019_CVPR}.
The second major issue of WSOD is \emph{grouped instances},
where 
neighbouring instances of objects in the same category are grouped into one large proposal, rather than proposed separately.
As image-level annotations reveal only the presence of each object class, without any information about object location or counts (see~\cref{fig:figure1}(b)), it has become conventional to take only the single highest-score proposal as a ``pseudo groundtruth'' \cite{Bilen_2016_CVPR,Tang_2017_CVPR,huang2020comprehensive}.
This can help avoid false positives,
but
often causes \emph{missing objects}, where less-obvious instances are ignored. 

Current argmax-based algorithms for finding pseudo groundtruths turn out to be problematic even on extremely popular benchmark datasets.
Labeling only one proposal per category misses $40\%$ of labels on PASCAL VOC~\cite{Everingham15} (selecting $7{,}306$ of $12{,}608$ target objects on VOC07),
and $60\%$ on MS-COCO~\cite{lin2014microsoft} ($533{,}396$ of $894{,}204$ objects on COCO14).
Similar patterns hold for VOC12 and COCO17 (see \cref{fig:figure1}(c)).
Object detection models trained with this limited and, indeed, potentially confusing supervision are substantially hindered,
and mining more pseudo-labels is problematic since they are likely to be false positives.

We introduce a novel multiple instance labeling method which addresses the limitations of current labeling methods in WSOD.
Our proposed object discovery module explores all proposed candidates using a similarity measure to the highest-scoring representation.
We further suggest a weakly supervised contrastive loss (WSCL) to set a reliable similarity threshold.
WSCL encourages a model to learn similar features for objects in the same class,
and to learn discriminative features for objects in different classes.
To make sure the model learn appropriate features, we provide a large number of positive and negative instances for WSCL through three feature augmentation methods suitable for WSOD.
This well-behaved embedding space allows the object discovery module to find more reliable pseudo groundtruths.
The resulting model then detects less-discriminative parts of target objects, misses fewer objects, and better distinguishes neighbouring object instances, as we will demonstrate experimentally in \cref{sec:experiments}.
As a result, the proposed approach beats state-of-the-art WSOD performance on both MS-COCO and PASCAL VOC by significant margins.

\section{Related Work}
\label{sec:related_work}
\subsection{Weakly Supervised Object Detection}
Bilen \etal~\cite{Bilen_2016_CVPR} introduced the first MIL-based end-to-end WSOD approach, known as WSDDN, which includes both classification and detection streams. 
Based on the MIL-based method, later works in WSOD have attempted to generate instance-level pseudo groundtruths in various ways.

\noindent{\textbf{Self-Supervised Pseudo Labeling Approach.}}
To use instance-level supervision, Tang~\etal~\cite{Tang_2017_CVPR} suggest Online Instance Classifier Refinement (OICR),
which alternates between training instance classifier and selecting the most representative candidates.
The online classifier rectifies initially detected instances through multiple stages and updates instance-level supervision determined by spatial relations. 
Tang~\etal~\cite{tang2018pcl} expand clusters from OICR to include adjacent proposals that belong to the same cluster.
Kosugi~\etal~\cite{Kosugi_2019_ICCV} devise an instance labeling method to find positive instances based on a context classification loss, and to avoid negative instances using spatial constraints.
Zeng~\etal~\cite{zeng2019wsod2} show that the bottom-up evidence, unlike top-down class confidence, helps to recognize class-agnostic object boundaries.
Chen~\etal~\cite{chen2020slv} propose a spatial likelihood voting (SLV) system to vote the bounding boxes with the highest likelihood in spatial dimension.
Huang~\etal~\cite{huang2020comprehensive} propose Comprehensive Attention Self-Distillation (CASD) that learns a balanced representation via input-wise and layer-wise feature learning.
CASD aggregate attention maps, generated by multiple transformations and extracted from different levels of feature maps.

\noindent{\textbf{Multiple Instance Approach.}}
Previous works focus on selecting valid pseudo groundtruths based on location information, but most still rely on the argmax labeling method which considers only one instance.
Some attempts have been made, however, to provide multiple pseudo groundtruths.
C-WSL~\cite{Gao_2018_ECCV} uses per-class object count annotations, which can help effectively separate grouped instances.
OIM~\cite{lin2020object} exploits an object mining method to find undiscovered objects by calculating Euclidean distance between the core instance and its surrounding boxes.
Ren~\etal~\cite{ren2020instance} propose not only Multiple Instance Self-Training (MIST) to generate top-k scored proposals as pseudo groundtruths, but also parametric dropblock to adversarially drop out discriminative parts.
Yin~\etal~\cite{yin2021instance} introduce feature bank to provide one more pseudo groundtruth using the top-similarity scored instance.
These algorithms \cite{lin2020object,ren2020instance,yin2021instance} effectively find multiple instances per class, but their methods largely depend on heuristics, rather than learning.
Our proposed method instead explores all possible pseudo groundtruths in a more reliable way, with learning guided by a contrastive loss.

\subsection{Contrastive Learning}
Contrastive losses have been successful in
unsupervised and self-supervised learning for image classification tasks~\cite{chen2020simple,khosla2020supervised}.
For object detection tasks, Xie~\etal~\cite{xie2021detco} and Sun~\etal~\cite{sun2021fsce} demonstrate learning good embedding features via contrastive learning successfully improves the generalization ability of an object detector. 
One important factor of this success is to use effective mining strategies for positive and negative samples, which accelerates convergence and enhances the generalization ability of a model.
In image classification tasks, these sample pairs are usually identified either by class labels~\cite{khosla2020supervised} if available, or pairing images with versions that have been randomly altered with methods such as cropping, color distortion, or Gaussian blur~\cite{chen2020simple}. 
Schroff~\etal~\cite{schroff2015facenet} introduce a hard positive and negative mining strategy
based on the distance between anchor and positive samples, with full supervision.
However, it is difficult to mine positive and negative samples in WSOD setting, which assumes no instance-level labels are available.
Therefore, we propose feature augmentations to sample positives and negatives for contrastive learning in the WSOD setting.
To the best of our knowledge, our method is the first approach to incorporate contrastive learning into WSOD tasks.
Our proposed weakly supervised contrastive loss guides a model to learn consistent feature representations for objects in the same class and discriminative representations for ones in different classes, through mining positive and negative samples and augmenting intermediate features.

\section{Background}

\label{sec:preliminary}
As with most state-of-the-art WSOD models, our approach is also based on the MIL head of WSDDN~\cite{Bilen_2016_CVPR}, followed by the refinement head suggested by OICR~\cite{Tang_2017_CVPR}.
In this section, we describe how MIL and refinement heads work.

\subsection{Feature Extractor}
Let a batch $B=\{I^n, R^n, Y^n\}_{n=1}^{N}$ contain an image $I^n$, proposals $R^n=\{r_1,\dots, r_{M^n}\}$ with $M^n$ proposals for the image $I^n$, and image-level labels $Y^n=[y^{n}_1, ..., y^{n}_C]\in \{0, 1\}^{C}$ where C is the number of classes.
Given an image $I^n$, a feature extractor generates features for downstream tasks as follows.
A backbone network takes a given image as input and outputs a feature map, from which a Region of Interest (RoI) feature map $f^n \in \mathbb{R}^{D \times H \times W \times M^n}$ is generated through an RoI pooling layer.
Two fully-connected (FC) layers, which we denote $\eta(\cdot)$, map $f^n$ to RoI feature vectors $v^n  \in\mathbb{R}^{D'\times M^n}$.
To alleviate part domination, we also randomly mask out some blocks of the RoI map with Dropblock \cite{ren2020instance}, generating $\tilde f^n$;
we then generate regularized feature vectors $\tilde v^n = \eta(\tilde f^n)$.
The MIL and refinement heads operate on $\tilde{v}^n$.

\begin{figure*}[t!]
\centering
\includegraphics[width=1.0\textwidth]{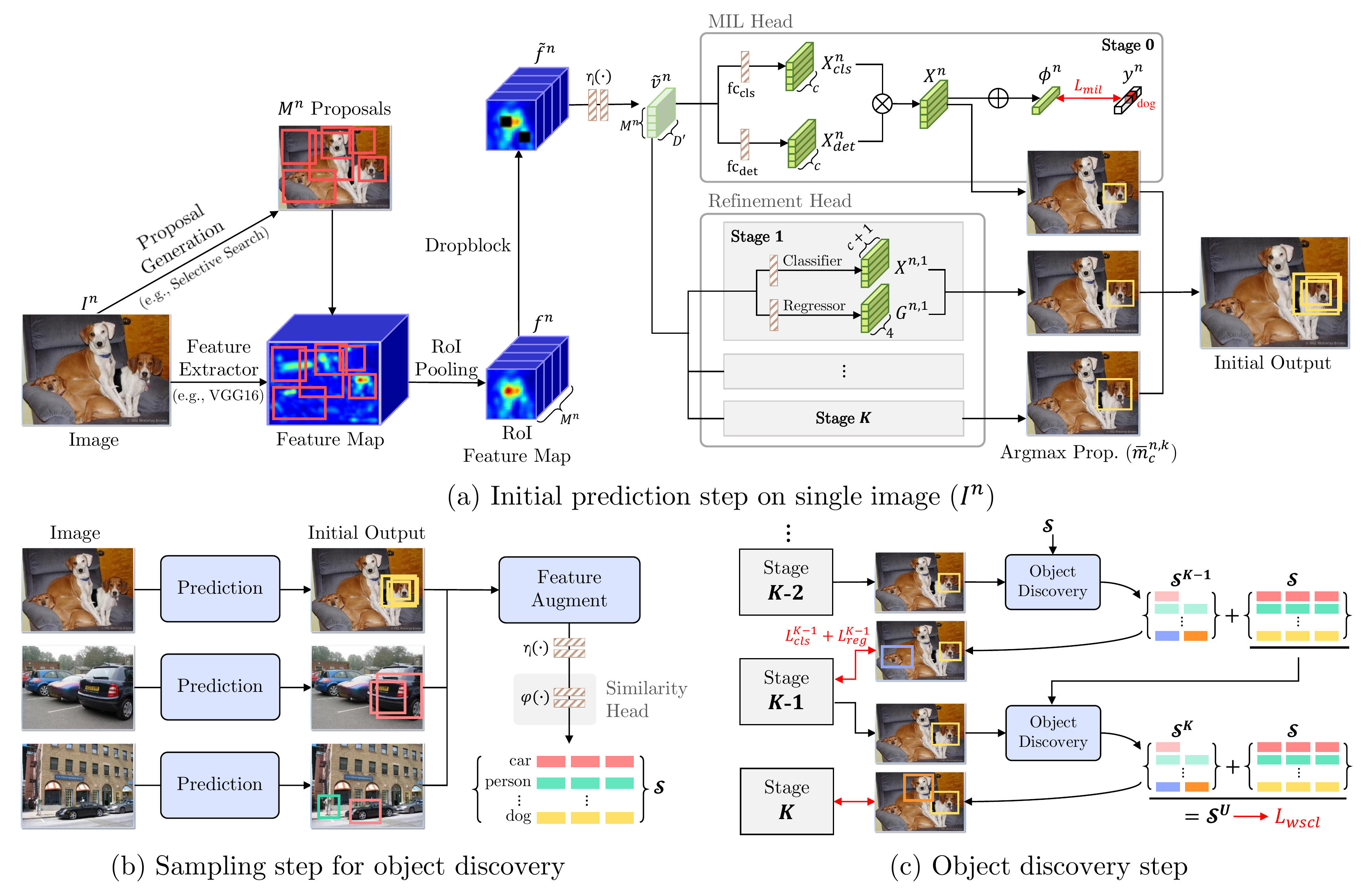}
\caption{Overall architecture of the proposed method. Initial prediction in (a) collects top-scoring instances over all stages. Sampling step for Object Discovery in (b) iterates step (a) for all images in a batch, and applies feature augmentations described in \cref{sec:sampling_strategy}.
Object discovery in (c) mines additional pseudo groundtruths that are not recognized by the argmax method.}
\label{fig:figure2_model}
\end{figure*}

\subsection{Multiple Instance Learning Head}
\label{sec:MIL}
As illustrated in \cref{fig:figure2_model}(a), Multiple Instance Learning (MIL) head consists of classification and detection networks which take RoI feature vectors $\tilde{v}^n$ as input, and return classification scores $X^{n}_{cls} \in\mathbb{R}^{C \times M^n}$ and detection scores $X^{n}_{det} \in\mathbb{R}^{C \times M^n}$. Here, the $X^{n}_{cls}$ are computed by a softmax operation along the classes (rows), whereas $X^{n}_{det}$ are computed along the regions (columns).
Proposal scores $X^{n} \in\mathbb{R}^{C\times M^n}$ are the element-wise product of classification and detection scores: $X^{n}=X^{n}_{cls} \odot X^{n}_{det}$.
The image score of the $c$-th class, $\phi^{n}_c$, is obtained by the sum of proposal scores over all regions: $\phi^{n}_c = \sum_{m=1}^{M^n} X^{n}_{c,m}$.
Given an image-level label $Y^n$ and image score $\phi^{n}_c$, the multi-label classification loss $L_{mil}$ is
  
\begin{equation}
\small
    L_{mil} = -\frac{1}{N}\sum_{n=1}^{N}\sum_{c=1}^{C} {y^{n}_c\log\phi^{n}_c + (1-y^{n}_c)\log(1-\phi^{n}_c)}
\label{eq:mil_loss}
.\end{equation}\par

\subsection{Refinement Head}
\label{sec:REF}
The goal of the refinement head is to integrate self-supervised training strategy into WSOD via instance-level supervision. 
At the $k$-th stage ($k \in \{1, ..., K\}$), an instance classifier generates proposal scores $X^{n,k} \in\mathbb{R}^{(C+1) \times M^n}$ where $M^n$ is the number of proposals and $C+1$ adds a background class to the $C$ classes.
Instance-level supervision at the $k$-th stage is determined by previous stage; in particular, the first instance classifier takes supervision from the output of MIL head.
Instance-level pseudo labels for the $n$-th image $Y^{n,k} \in\mathbb{R}^{(C+1) \times M^n}$ are then set to $1$ if the corresponding proposal sufficiently overlaps the highest scored proposal, otherwise $0$ as defined in \eqref{eq:argmax_and_adjacents}.
\begin{equation}
    \bar{m}^{n,k}_c = \argmax_m  {x^{n,(k-1)}_{c,m}};
    \qquad
    y^{n,k}_{c,m} =
      \begin{cases}
        1  & \text{if $IoU(r_m, r_{\bar{m}^{n,k}_c}) > 0.5$}\\
        0  & \text{otherwise.}
      \end{cases}
\label{eq:argmax_and_adjacents}
\end{equation}
Finally, the instance classification loss $L_{cls}$ is defined as
\begin{equation}
      L_{cls} = -\frac{1}{N}\sum_{n=1}^{N}\frac{1}{K}\sum_{k=1}^{K}\frac{1}{M^n}\sum_{m=1}^{M^n}\sum_{c=1}^{C+1} w^{n,k}_m y^{n,k}_{c,m}\log{x^{n,k}_{c,m}}
\label{eq:b_cls_loss}
\end{equation}
where $x^{n,k}_{c,m}$ denotes $m$-{th} proposal score of a class $c$ at $k$-th stage, $w^{n,k}_m$ denotes a loss weight defined as $w^{n,k}_m = x^{n,(k-1)}_{c,\bar{m}^{n,k}_c}$ following OICR~\cite{Tang_2017_CVPR}, and $K$ is the total number of refinement stages.
  
In addition to instance classification, some work \cite{zeng2019wsod2,Yang_2019_ICCV} has improved localization performance by adding a bounding box regression loss.
Given $\hat M^n$ pseudo groundtruth bounding boxes $\hat g^{n,k}_m$,
nearby predicted bounding boxes $g^{n,k}_m$, matched as in \eqref{eq:argmax_and_adjacents},
are encouraged to align using a $smooth_{L1}$ regression loss:
\begin{equation}
      L_{reg} = -\frac{1}{N}\sum_{n=1}^{N}\frac{1}{K}\sum_{k=1}^{K}\frac{1}{\hat{M}^n}\sum_{m=1}^{\hat{M}^n} w^{n,k}_m smooth_{L1}(g^{n,k}_m, \hat{g}^{n,k}_m)
\label{eq:b_reg_loss}
.\end{equation}

\section{Our Approach}
\label{sec:approach}
Most prior WSOD works~\cite{Tang_2017_CVPR,huang2020comprehensive} consider only top-scoring proposals, as described in \eqref{eq:argmax_and_adjacents}.
This strategy, however, has significant challenges in achieving our goal of detecting \emph{all} objects present.
To alleviate this issue, we propose a novel approach called \textit{object discovery} which secures reliable pseudo groundtruths, by transferring instance-level supervision from previous to the next stage as illustrated in \cref{sec:REF} and \cref{fig:figure2_model}(c).
To further enhance the object discovery module, we also introduce a new \textit{similarity head} which maps RoI feature vectors to an embedding space, guided by a novel weakly supervised contrastive loss (WSCL).


\subsection{Similarity Head}
Parallel to the MIL and refinement heads, we construct a similarity head $\varphi{(\cdot)}$ that takes augmented RoI feature vectors as input described in \cref{fig:figure2_model}.
We will explain how RoI features are augmented in \cref{sec:sampling_strategy}.
The similarity head consists of two FC layers which map the inputs to a 128-dimensional space, followed by a normalization step.
Thus, the outputs of the similarity head are expressed as $z^n = \varphi{(v^n)} \in\mathbb{R}^{128 \times M^n}$ where $||z^n_m||_2 = 1$.
Note that the similarity head uses $v^n$, whereas MIL and refinement head use the region-dropped $\tilde{v}^n$.

\subsection{Sampling Strategy for Object Discovery}
\label{sec:sampling_strategy}

\begin{figure}[t!]
\centering
\includegraphics[width=1.0\columnwidth]{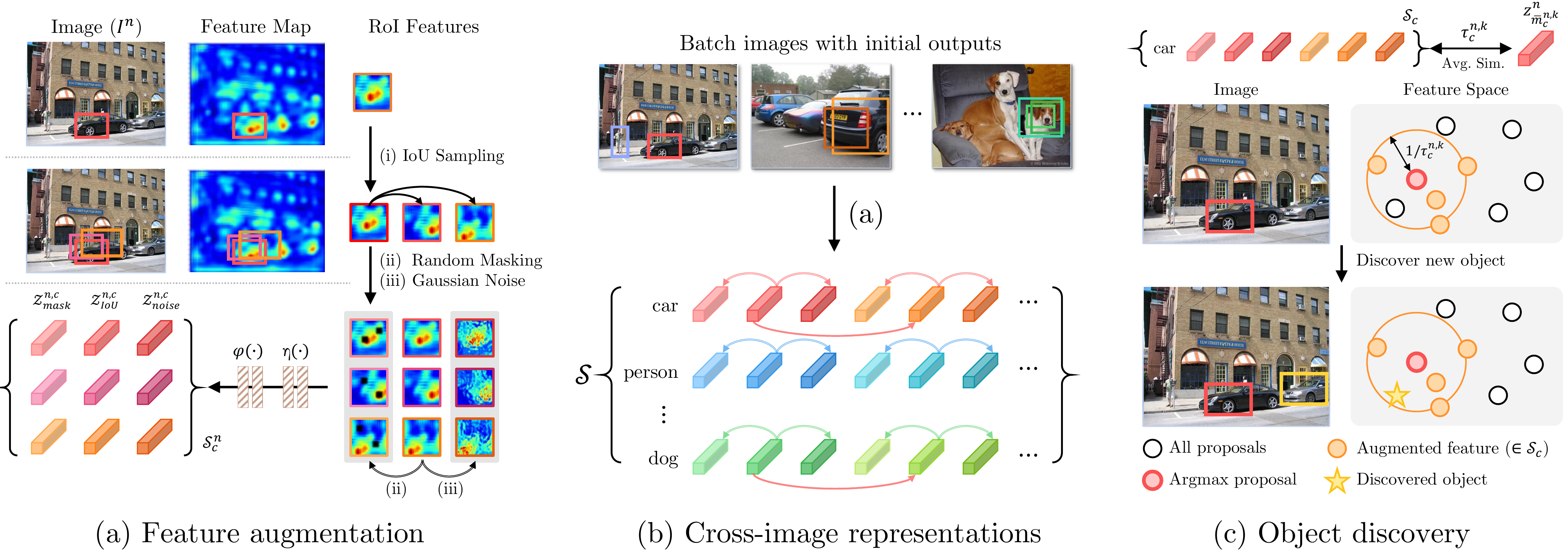}
\caption{Three steps for object discovery module.
(a) applies feature augmentation methods to embedding vectors of top-scoring proposals.
(b) collects all augmented embedding vectors through all images in a batch.
(c) determines new pseudo groundtruths based on the similarity with the embedding vector of the top-scoring instance $z^{n}_{\bar{m}^{n,k}_c}$ and similarity threshold $\tau^{n,k}_c$.}
\label{fig:figure3_details}
\end{figure}

Contrastive learning \cite{khosla2020supervised,chen2020simple}, in general, focuses on making ``positive'' and ``negative'' pairs have similar and different feature embeddings, respectively.
Although it is possible to augment images and pass each to the backbone to obtain pair of samples from RoI features in WSOD setting where no instance-level supervisions are available, it is computationally inefficient: most of the features from the backbone are not used as RoI features. 
Instead of augmenting images, we propose three feature augmentation methods to generate views of samples, as shown in
\cref{fig:figure3_details}(a):
\textit{IoU sampling}, \textit{random masking}, and \textit{adding gaussian noise}.

\noindent\textbf{IoU Sampling.}
The purpose of IoU sampling is to increase the number of samples by treating the proposals adjacent to the top-scoring proposal $\bar{m}^n_c$ in \eqref{eq:argmax_and_adjacents} as positives. 
The proposals that overlap more than a threshold $\tau_{IoU}$ with the top-scoring proposal at each stage $k$ are considered positive samples, and the corresponding embedding vectors are formulated as
\begin{equation}
\begin{gathered}
    \mathcal{M}^{n,k}_c = \{m \mid IoU(r_m, r_{\bar{m}^{n,k}_c}) > \tau_{IoU}, m = 1, 2, ..., M^n \}\\
    \mathcal{Z}^{n,c}_{IoU} = \{ \varphi(\eta{(f^n_m)}) \ | \ m \in \bigcup_{k=0}^{K-1} \mathcal{M}^{n,k}_c \}
\label{eq:IoU_sampling}
\end{gathered}
\end{equation}
where $M^n$ denotes the total number of proposals in $n$-th image, $\eta(\cdot)$ is the extractor of RoI feature vectors, and $\varphi(\cdot)$ denotes the similarity head.

\noindent\textbf{Random Masking.}
Random masking randomly drops some regions across all channels of a RoI feature map.
We first generate a random map $D: D_{i,j}\sim U(0,1) \in \mathbb{R}^{H \times W}$.
Then the binary mask $D_{drop}$ is determined by drop threshold $\tau_{drop}$, so if $D < \tau_{drop}$, $D_{drop}$ is set to be 0, otherwise 1.
Finally, a randomly-masked feature is obtained by taking spatial-wise multiplication of a RoI feature map $f^n_m$ with $D_{drop}$ followed by the similarity head,
\begin{equation}
    \mathcal{Z}^{n,c}_{mask} = \{ \varphi(\eta{(f^n_m \odot D_{drop})}) \ | \ m \in \bigcup_{k=0}^{K-1} \mathcal{M}^{n,k}_c \}
\label{eq:mask_sampling}
.\end{equation}
Here, random masking is applied to the all proposals from IoU sampling at all stages, $\bigcup_{k=1}^K \mathcal{M}^{n,k}_c$, to obtain more positive samples.

\noindent\textbf{Adding Gaussian Noise.}
To add Gaussian random noise to RoI feature maps, we create a random noise map $D_{noise}: D_{i,j} \sim N(0,1) \in \mathbb{R}^{H \times W}$.
We add this to the RoI feature maps $f^n_m$ by
\begin{equation}
    \mathcal{Z}^{n,c}_{noise} = \{ \varphi(\eta(f^n_m + f^n_m \odot D_{noise})) \ | \ m \in \bigcup_{k=0}^{K-1} \mathcal{M}^{n,k}_c \}
\label{eq:noise_sampling}
.\end{equation}


\noindent\textbf{Cross-Image Representations.}
Finally, we gather the augmented embedding vectors corresponding to the same object category from different images, as described in \cref{fig:figure3_details}(b).
We treat cross-batch representations from the same categories as positive examples in the mini-batch,
\begin{equation}
    \mathcal{S}_c = \bigcup_{n=1}^N ( \mathcal{Z}^{n,c}_{IoU} \cup \mathcal{Z}^{n,c}_{mask} \cup \mathcal{Z}^{n,c}_{noise} )
\label{eq:cross}
.\end{equation}

\subsection{Object Discovery}
Using the augmented RoI features introduced in the previous section, we discover many reliable instance-level pseudo groundtruths missed by previous methods that take only the top-scoring proposals.
Intuitively, even though the classification score of a proposal may be low, if its embedding vector is close to that of the top-scoring proposal, it is likely that the proposal shares the same class as the top-scoring proposal.
Therefore, instead of solely relying on classification scores, we exploit similarity scores between the embedding vectors of all proposals and the top-scoring (argmax) proposal at each stage $k$, to discover additional pseudo grountruths as shown in \cref{fig:figure3_details}(c).
To mine new pseudo groundtruths, we first compute a threshold $\tau^{n,k}_c$ that determines whether to label a proposal as a pseudo groundtruth for a class $c$ at stage $k$:
\begin{equation}
    \tau^{n,k}_c = \frac{1}{|\mathcal{S}_c|}\sum^{|\mathcal{S}_c|}_{i=1}{sim(z^{n}_{\bar{m}^{n,k}_c}, \mathcal{S}_{c,i})}
\label{eq:object_discovery_threshold}
,\end{equation}
where $z^{n}_{\bar{m}^{n,k}_c}$ denotes embedding vectors of top-scoring proposals at stage $k$, $S_{c,i}$ denotes $i$-th element of $S_c$, and $sim(\cdot,\cdot)$ is a dot product between inputs.
Then, new pseudo groundtruth candidates $\acute{\mathcal{M}}^{n,k}_c$ are determined as the ones having higher similarity to the top-scoring proposal than similarity threshold $\tau^{n,k}_c$, 
\begin{equation}    
    \acute{\mathcal{M}}^{n,k}_c = \{m \mid sim(z^{n}_m, z^{n}_{\bar{m}^{n,k}_c}) > \tau^{n,k}_c, m=1, 2, ...,M^n \}.
    \label{eq:PGT_discovery}
\end{equation}
Finally, we add new pseudo groundtruths denoted as $\tilde{\mathcal{M}}^{n,k}_c$ at stage k after applying Non-Maximum Suppression (NMS)~\cite{rosenfeld1971edge} to $\acute{\mathcal{M}}^{n,k}_c$. 

Consequently, we update instance-level supervision and embedding vectors for newly discovered pseudo groundtruths $\tilde{\mathcal{M}}^{n,k}_c$.
We re-label instance-level supervision $\{ y^{n,k}_{c,m} | m \in \tilde{\mathcal{M}}^{n,k}_c \}$ and its adjacent proposals as described in \eqref{eq:argmax_and_adjacents}.
New embedding vectors $\{ z^{n}_{m} | m \in \tilde{\mathcal{M}}^{n,k}_c \}$ are add to $S^k_c$, as it is expected discriminative features help to measure precise similarity.
Then, classification loss $L_{cls}$ in \eqref{eq:b_cls_loss} and regression loss $L_{reg}$ in \eqref{eq:b_reg_loss} are updated accordingly.
After going through all the stages from $1$ to $K$, we obtain $S^U = S^K \cup S$ as described in \cref{fig:figure2_model}(c).

\subsection{Weakly Supervised Contrastive Loss}

To learn more consistent feature representations for the proposals in the same class, we propose weakly supervised contrastive loss (WSCL) that learns representations by attracting positive samples closer together and repelling negative samples away from positives samples in the embedding space.
From a collection $S^U$ = $\{s_i, t_i\}_{i=1}^{|S^U|}$ where $s_i$ denotes $i$-th embedding vectors, $t_i$ denotes the pseudo label of $s_i$, WSCL for $i$-th embedding vector denoted as $L^{i}_{wscl}$, is formulated as
\begin{equation}
  L^{i}_{wscl} = -\frac{1}{N_{t_i}-1}\sum_{j=1,j\neq i}^{|S^U|}{\mathbbm{1}\{t_i=t_j\}}\cdot\log{\frac{\exp{(s_i\cdot s_j/\epsilon)}}{\sum_{l=1,l\neq i}^{|S^U|} \exp{(s_i\cdot s_l/ \epsilon)}}}\\
\label{eq:wscl_i}
\end{equation}
where $N_{t_i} \coloneqq \sum_{j=1}^{|S^U|}{\mathbbm{1}\{t_i=t_j\}} $, and $\epsilon$ is a temperature parameter introduced in~\cite{khosla2020supervised}.
Note that $\mathcal{S}^U = \bigcup_{c=1}^C \mathcal{S}^U_c$.

\noindent\textbf{Instance difficulty.}
Since confidence score of instance is noisy at early stages of training, we introduce instance difficulty $\omega$ to make training for WSCL easier.
$\omega$ is the set of scores for all images in a batch where each score is the instance score from the MIL head over the sum of them at each image.
Here, the size of $\omega$ is the same as $\mathcal{S}$.
Then, the re-weighted contrastive loss is formulated as,



\begin{equation}
   L_{wscl} = \frac{1}{|S^U|}\sum_{i=1}^{|S^U|} \omega_{i}\cdot L^{i}_{wscl}, \ \omega = \bigcup_{n=1}^N \left\{ {X^n_{c,m}} / \sum_{j=1}^{M^n} X^{n}_{c,j} \ | \ m \in \bigcup_{k=0}^{K-1} \mathcal{M}^{n,k}_c \right\}
\label{eq:wscl}
\end{equation}

\noindent\textbf{Total Loss.}
Finally, the total loss of training the proposed model is defined as
\begin{equation}
  L_{total} = L_{mil} + L_{cls} + L_{reg} + \lambda L_{wscl}
\label{eq:tot_loss}
\end{equation}
where $\lambda$ is a loss weight to balance scale with the other losses.

\section{Experiments}

\subsection{Experiment Setting}
\label{sec:experiments}
\noindent\textbf{Datasets.}
To verify the robustness of our method, we evaluate it on four object detection datasets; VOC07 and VOC12 in PASCAL VOC~\cite{Everingham15}, and COCO14 and COCO17 in MS-COCO~\cite{lin2014microsoft}, following the convention in WSOD tasks.
We use trainval sets containing 5,011 and 11,540 images for VOC07 and VOC12, respectively, and test sets that contain 4,951 and 10,991 images, for evaluation.
We further investigate the robustness of our method on MS-COCO datasets.
For COCO14, we train our model on the train set of 82,783 images and test it with the validation set of 40,504 images.
For COCO17, we split the dataset into the train set of 118,287 images and validation set of 5,000 images.
We use only image-level annotations to train our model on all datasets.


\noindent\textbf{Evaluation Metrics.}
On VOC07 and 12 datasets, we evaluate our model on the test set using mean Average Precision (mAP) metric with standard IoU criterion (0.5).
MS-COCO is more challenging than PASCAL VOC as it has significantly more instances per image (about $2$ \vs{} $7$) and more classes ($20$ \vs{} $80$). 
For this reason, MS-COCO is often not considered in the WSOD literature.
We report the performance on MS-COCO datasets following the standard COCO metric which includes several metrics, such as, average precision (AP) and average recall (AR) with varying IoU thresholds \eg 0.5 and 0.75, and object sizes \eg small (\textit{s}), medium (\textit{m}), and large (\textit{l}), but the most representative metric is the AP averaged over 10 IoU thresholds (from 0.5 to 0.95 for every 0.05 step). 

\noindent\textbf{Implementation Details.}
All the implementation is developed in PyTorch.
For both VGG16~\cite{VGG:Simonyan2014} and ResNet~\cite{he2016deep} models, we initialize parameters using ImageNet~\cite{deng2009imagenet} pre-trained networks.
For VGG16, following the previous methods~\cite{ren2020instance,huang2020comprehensive}, we replace a global average pooling layer with a RoI pooling layer, and remove the last FC layer leaving two FC layers, which all the heads including the similarity head are attached to.
For ResNet, we modify the structure for WSOD as suggested in Section 4 of Shen~\etal~\cite{shen2020enabling}.
We use around $2{,}000$ proposals per image for both proposal generation methods, SS~\cite{uijlings2013selective} and MCG~\cite{arbelaez2014multiscale}.

\noindent\textbf{Hyperparameters.}
The batch size is set to $8$ for PASCAL VOC and MS-COCO datasets. We train models for 30K, 60K and 130K iterations on VOC07, VOC12 and MS-COCO, respectively, using a SGD optimizer with the learning rate of $0.01$ and weight decay of $0.0001$ except for R50-WS and R101-WS~\cite{shen2020enabling} where the learning rate is set to $0.02$ on both datasets.
At inference time, the prediction scores are computed as the average of proposal scores for all $k$-stages, and the offsets from regression branch are incorporated to adjust the coordinates of bounding boxes.
The final predictions are made after applying NMS of which threshold is set to 0.4 for both datasets.
Following the previous methods~\cite{ren2020instance,Tang_2017_CVPR,huang2020comprehensive}, the inputs are multi-scaled to $\{480, 576, 688, 864, 1000, 1200\}$ for both training and inference time.
In the experiments, we set $\tau_{IoU} = 0.5$, $\tau_{drop} = 0.3$, $\tau_{nms} = 0.1$, $\lambda = 0.03$ ($\lambda = 0.01$ on COCO datasets) and $\epsilon = 0.2$ for WSCL and K=3 for the number of refinement stages. But, as we will show in an ablation study, the performance is not sensitive to the choice of hyperparameters.

\begin{table*}[t!]
\caption{Comparison of the state-of-the-art algorithms on MS-COCO}
\centering
\setlength{\tabcolsep}{4.0pt}
\resizebox{\textwidth}{!}{
\begin{tabular}{c|c|c|c|cc|ccc|ccc|ccc}
 \hline
 Dataset &Backbone &Method &$AP$ &$AP^{50}$ &$AP^{75}$ &$AP^s$ &$AP^m$ &$AP^l$ &$AR^1$ &$AR^{10}$ &$AR^{100}$ &$AR^s$ &$AR^m$ &$AR^l$\\
 \hline\hline
 \multirow{11}{*}{COCO14} & \multirow{6}{*}{VGG16}  &PCL~\cite{tang2018pcl}    &8.5 &19.4 &-&-&-&-&-&-&-&-&-&-\\
 &               &C-MIDN~\cite{gao2019c} &9.6 &21.4 &-&-&-&-&-&-&-&-&-&-\\
 &               &WSOD2~\cite{zeng2019wsod2}  &10.8 &22.7 &-&-&-&-&-&-&-&-&-&-\\
 &               &MIST~\cite{ren2020instance}   &11.4 &24.3 &9.4 &3.6 &12.2 &17.6 &13.5 &22.6 &23.9 &8.5 &25.4 &38.3 \\
 &               &CASD~\cite{huang2020comprehensive}   &12.8 &26.4 &-&-&-&-&-&-&-&-&-&-\\
 &               &Ours   &\textbf{13.7} &\textbf{27.7} &\textbf{11.9} &\textbf{4.4} &\textbf{14.5} &\textbf{21.2} &\textbf{14.7} &\textbf{24.8} &\textbf{26.9} &\textbf{8.8} &\textbf{27.8} &\textbf{44.0}  \\
\cline{2-15}
 & \multirow{3}{*}{ResNet50} &MIST~\cite{ren2020instance} &12.6 &26.1 &10.8 &3.7 &13.3 &19.9 &14.8 &23.7 &24.7 &8.4 &25.1 &41.8 \\
 &       &CASD~\cite{huang2020comprehensive}  &\textbf{13.9}      &27.8     &-           &-   &-&-&-&-&-&-&-&- \\
 &       &Ours
 &\textbf{13.9} &\textbf{29.1} &\textbf{11.8} &\textbf{4.9} &\textbf{16.8} &\textbf{22.3} &\textbf{15.5} &\textbf{26.1} &\textbf{28.0} &\textbf{9.0} &\textbf{31.8} &\textbf{46.6} \\
 \cline{2-15}
 &\multirow{2}{*}{ResNet101}      &MIST~\cite{ren2020instance} 
 &13.0      &26.1     &10.8         &3.7   &13.3 &19.9 &14.8 &23.7 &24.7 &8.4 &25.1 &41.8 \\
 &       &Ours     
 &\textbf{14.4} &\textbf{29.0} &\textbf{12.4} &\textbf{4.8} &\textbf{17.3} &\textbf{23.8} &\textbf{15.8} &\textbf{27.0} &\textbf{30.0} &\textbf{9.2} &\textbf{33.6} &\textbf{51.0}\\
 \hline
 \multirow{4}{*}{COCO17}  &\multirow{2}{*}{VGG16}      &MIST~\cite{ren2020instance} 
 &12.4      &25.8     &10.5        &3.9  &13.8 &19.9 &14.3 &23.3 &24.6 &\textbf{9.7} &26.6 &39.6 \\
 &                              &Ours  
 &\textbf{13.6} &\textbf{27.4} &\textbf{12.2} &\textbf{4.9} &\textbf{15.5} &\textbf{21.6} &\textbf{14.6} &\textbf{24.8} &\textbf{26.8} &9.2 &\textbf{28.7} &\textbf{43.8} \\
\cline{2-15}
 &ResNet50       &Ours &13.8 &27.8 &12.1 &5.7 &17.7 &23.8 &15.1 &26.6 &29.7 &10.1 &33.7 &50.7\\
\cline{2-15}
 &ResNet101      &Ours &14.4 &28.7 &12.6 &5.4 &17.9 &25.5 &15.4 &26.8 &29.6 &10.0 &33.3 &50.6\\
 \Xhline{2\arrayrulewidth}
\end{tabular}}
\label{tbl:sota_coco}
\end{table*}

\subsection{Quantitative Results}
\noindent\textbf{Comparison with state-of-the-arts.}
In \cref{tbl:sota_coco}, we compare the proposed
method with other state-of-the-art algorithms on COCO14 and 17.
Regardless of backbone structure and dataset, our method achieves the new state-of-the-art performance for all the evaluation metrics.
The fact that the performance of Ours in AR measurements are higher than the other methods implies that the proposed method successfully detects missing instances compared to the previous methods on MS-COCO, which contain many instances
and categories per image.
For instance, our method outperforms MIST~\cite{ren2020instance} by a large margin in AR measurements; on average of $2.13\%$ for $(AR^1, AR^{10}, AR^{100})$ 
and on average of $3.46\%$ for $(AR^s, AR^m, AR^l)$ with VGG16 on COCO14.
Despite the significant improvement in AR, AP also improves by a large margin regardless of different sub-category of AP and backbone structure.
Our method gains on average of $2.95\%$ for $(AP^{50}, AP^{75})$ and on average of $2.23\%$ for $(AP^s, AP^m, AP^l)$ with VGG16 on COCO14.
We observe similar tendency on COCO17 for all the backbones.
As a result, our proposed method achieves the new state-of-the-art performance
\begin{wraptable}[21]{r}{0.45\textwidth}
\caption{Comparison of the state-of-the-art methods on PASCAL VOC.}
    \centering
    \fontsize{8.5}{10.5}\selectfont
    \begin{tabu}{c|c|c|c}
        \hline
        Proposal & Method & VOC07 &VOC12\\
        \hline\hline
        \multirow{13}{*}{SS\cite{uijlings2013selective}} & WSDDN\cite{Bilen_2016_CVPR}      &34.8 &- \\
        & OICR\cite{Tang_2017_CVPR}        &41.2 &37.9 \\
        & PCL\cite{tang2018pcl}            &43.5 &40.6 \\
        & C-WSL\cite{Gao_2018_ECCV}        &46.8 &43.0 \\
        & WSRPN\cite{Tang_2018_ECCV}       &47.9 &43.4 \\
        & C-MIL\cite{Wan_2019_CVPR}        &50.5 &46.7 \\
        & C-MIDN\cite{gao2019c}            &52.6 &50.2 \\
        & WSOD2\cite{zeng2019wsod2}        &53.6 &47.2 \\
        & OIM\cite{lin2020object}          &50.1 &45.3 \\
        & SLV\cite{chen2020slv}            &53.5 &49.2 \\
        & MIST\cite{ren2020instance}       &54.9 &52.1 \\
        & CASD\cite{hwang2021weakly}       &\textbf{56.8} &53.6 \\
        & Ours                             &56.1 &\textbf{54.6} \\
        \hline
        \multirow{3}{*}{MCG\cite{arbelaez2014multiscale}} & MIST\cite{ren2020instance}  &56.5 &53.9\\
        & CASD\cite{hwang2021weakly}       &57.4 &-\\
        & Ours & \textbf{58.7} & \textbf{56.2} \\
        \Xhline{2\arrayrulewidth}
    \end{tabu}
\label{tbl:sota_voc}
\end{wraptable}
on both COCO14 and COCO17 ($14.4\%$).
In \cref{tbl:sota_voc}, we compare the performance of the state-of-the-art methods on PASCAL VOC with both SS and MCG proposal methods.
Since PASCAL VOC datasets contain less number of instances and categores per image, the gain achieved by our method is relatively lower than MS-COCO.
It, however, outperforms other multiple instance labeling methods~\cite{Gao_2018_ECCV,lin2020object,ren2020instance} with clear margins.
For example, Ours outperforms MIST~\cite{ren2020instance} (the best performance multiple instance labeling method) by $1.2\%$ and $2.5\%$ on VOC07 and VOC12, respectively, with SS proposal method, and $2.2\%$ and $2.3\%$ with MCG proposal method.
It also achieves the new state-of-the-art performance on VOC12 ($54.6\%$) and compatible results on VOC07 (Ours: $56.1\%$ \vs{} CASD: $56.8\%$).
We use MCG for the rest of the experiments as it outperforms SS.

\newcolumntype{C}{>{\centering\arraybackslash}p{3em}}
\newcolumntype{D}{>{\centering\arraybackslash}p{2.4em}}
\begin{table}[t!]
\caption{Experiment results with various settings on VOC07. (a) Different components of the proposed method. (b) Performance with ResNet backbones~\cite{shen2020enabling}. (c) Comparison of different combination of feature augmentation methods.}
    \scriptsize{
    \begin{subtable}{.28\textwidth}
    \captionsetup{font=small}
        \centering
        \begin{tabular}[t]{CC|C}
            \firsthline
            OD  & WSCL & mAP\\ 
            \hline\hline
            &  & 52.3 \\
            \checkmark & & 56.1 \\
            & \checkmark & 54.5 \\
            \checkmark & \checkmark & \textbf{58.7} \\
            \Xhline{2\arrayrulewidth}
        \end{tabular}
            \caption{Diff. components}
        \label{tbl:OD/WSCL}
    \end{subtable}}
    \scriptsize{
    \begin{subtable}{.3\textwidth}
    \captionsetup{font=small}
        \centering
        \begin{tabular}[t]{c|c|c}
            \firsthline
            Method &R50-WS &R101-WS\\
            \hline\hline
            OICR\cite{Tang_2017_CVPR}     &50.9 &51.4\\
            PCL\cite{tang2018pcl}         &50.8 &53.3\\
            C-MIL\cite{Wan_2019_CVPR}     &53.4 &53.9\\
            Ours                          &\textbf{56.6} &\textbf{56.5}\\
            \Xhline{2\arrayrulewidth}
        \end{tabular}
        \caption{ResNet backbones}
        \label{tbl:backbone_resnet}
    \end{subtable}}\hfill
    \scriptsize{
    \begin{subtable}{.38\textwidth}
    \captionsetup{font=small}
        \centering
        \begin{tabular}[t!]{c|DDD|c}
            \firsthline
            Method   &IoU &Mask &Noise &mAP\\
            \hline\hline
            OICR$^+$ &  & &  &52.3 \\
            \hline
            \multirow{7}{*}{+ Ours}
            &\checkmark &  &               &56.8\\
            &  &\checkmark &               &55.1\\ 
            &  &  &\checkmark              &54.8\\
            &\checkmark &\checkmark &      &57.4\\
            &\checkmark &  &\checkmark     &57.4\\
            &  &\checkmark &\checkmark     &55.2\\
            &\checkmark &\checkmark &\checkmark &\textbf{58.7}\\
            \Xhline{2\arrayrulewidth}
        \end{tabular}
        \caption{Feature augmentation}
        \label{tab:components}
     \end{subtable}}
\label{tbl:table3}
\end{table}
\noindent\textbf{The effectiveness of each component.}
To validate the effectiveness of each component of Object Discovery (OD) and WSCL modules, we provide experiment results on VOC07 with SS in \cref{tbl:table3}(a).
Note that we find the performance of OICR~\cite{Tang_2017_CVPR} can be further increased by adding the bounding box regression and dropblock~\cite{ghiasi2018dropblock} layers ($45.0\% \rightarrow 52.3\%$ on VOC07), thus, we call OICR + Regression + Dropblock as OICR$^+$, and use it as the baseline throughout the experiments unless specified otherwise.
In \cref{tbl:table3}(a), each of OD and WSCL modules significantly improves OICR$^+$ baseline ($+3.8$ and $+2.2$) but the improvement is the highest when both OD and WSCL are applied simultaneously, which partially demonstrate that each module helps the other one as described in \cref{sec:approach}.

\noindent\textbf{Robust regardless of backbone.}
In \cref{tbl:sota_coco}, we have shown that the proposed method performs well both in AP and AR metrics on MS-COCO datasets regardless of backbone structure. 
Although it is not a common practice to provide the performance with ResNet backbones on PASCAL VOC, we provide the performance of our method on ResNet in \cref{tbl:table3}(b).
It shows that Ours with ResNet backbones significantly outperforms the previous state-of-the-art methods as with the case with VGG backbone shown in \cref{tbl:sota_voc}.
The performance of the previous methods are reported in~\cite{shen2020enabling}.

\subsection{Ablation Studies}
\noindent\textbf{Feature augmentation methods.}
To further verify the effectiveness of the proposed feature augmentation methods, we experiment with different combination of augmentation methods for the object discovery and WSCL modules in \cref{tbl:table3}(c).
The performance significantly improves from $52.3\%$ to $55.1\%$ and $54.8\%$ with random masking and Gaussian noise, respectively, even without IoU sampling.
With IoU sampling prior to random masking and Gaussian noise, the performance consistently improves further by $2.3\%$ and $2.6\%$ for random masking and Gaussian noise.
Using all the proposed feature augmentations, the performance reaches $58.7\%$ that is $6.4\%$ higher than the baseline.

\begin{figure}[t!]
\centering
\includegraphics[width=1.0\columnwidth]{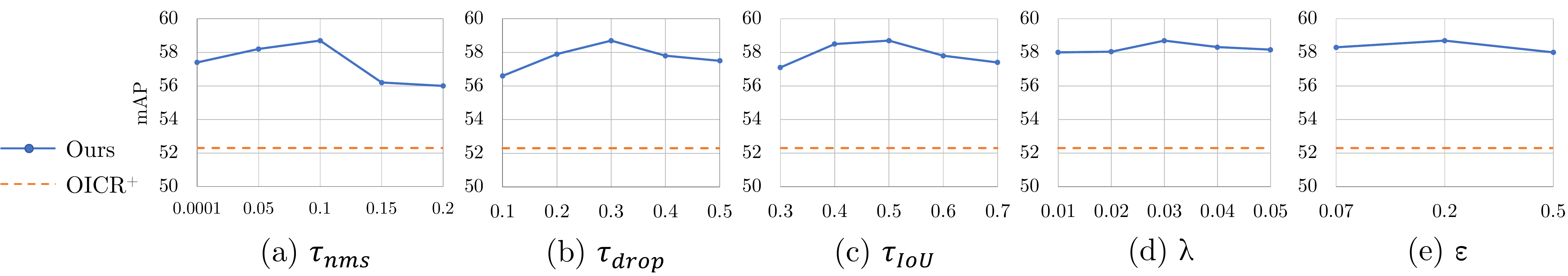}
\caption{Performance with different values of $\tau_{nms}$, $\tau_{drop}$, $\tau_{IoU}$, $\lambda$ and $\epsilon$.}
\label{fig:abl_study}
\end{figure}

\noindent\textbf{Sensitivity to hyperparameters.}
In \cref{fig:abl_study}, we provide the experiment results with different values of the hyperparameters we introduce.
Regardless of hyperparameter, the performance is not sensitive to the choice of values around the optimal values we choose ($\tau_{nms}=0.1, \tau_{drop}=0.3, \tau_{IoU}=0.5, \lambda=0.03$ and $\epsilon=0.2$).  
For instance, the gap between the highest and lowest performance for each hyperparameter is no more than $2.6\%$ in mAP (highest with $\tau_{nms}$), which demonstrates that our proposed method does not greatly depend on hyperparameter tuning.
In (e), we use the same values of $\epsilon$ following the experiments conducted in other contrastive learning methods~\cite{khosla2020supervised,chen2020simple}.

\begin{figure*}[t!]
\centering
\includegraphics[width=1.0\textwidth]{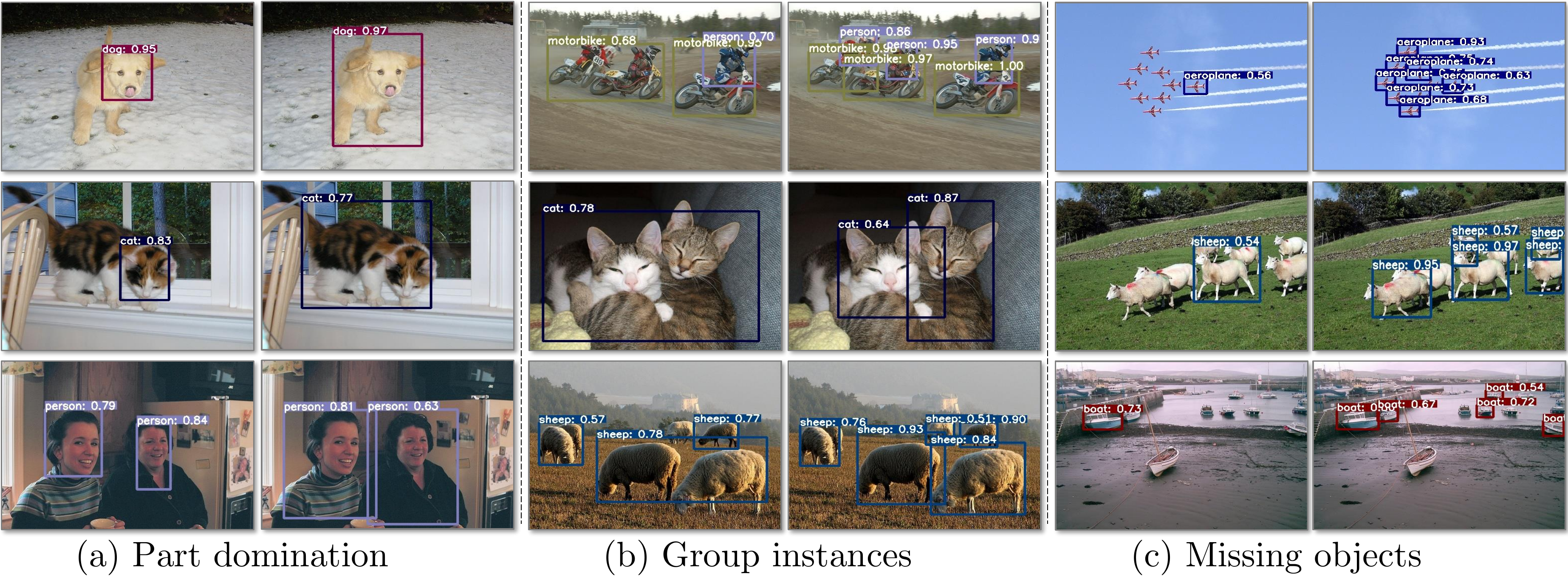}
\caption{Qualitative results of OICR~\cite{Tang_2017_CVPR} and ours about the three challenges of WSOD: (a) part domination, (b) grouped instances and (c) missing objects. 
The images on the left and right  indicate OICR and Ours, respectively.}
\label{fig:visualization}
\end{figure*}
\subsection{Qualitative results}
\noindent\textbf{Three challenges of WSOD.} In \cref{fig:visualization}, we provide the qualitative results that show how our method addresses three main challenges of WSOD -- part domination, grouped instances and missing objects (described in \cref{sec:intro}), compared to OICR~\cite{Tang_2017_CVPR}.
The left columns show the results from OICR~\cite{Tang_2017_CVPR} whereas the right columns show the results from our method.
The part domination shown in (a) is largely alleviated, especially for the categories with various poses such as \textit{dog}, \textit{cat} and \textit{person}.
We also observe that grouped instances are separated into multiple bounding boxes in (b).
Lastly, our method successfully detects many of instances that are ignored with the argmax labeling method as shown in (c).

\begin{figure}[t]
\centering
\includegraphics[width=1.0\columnwidth]{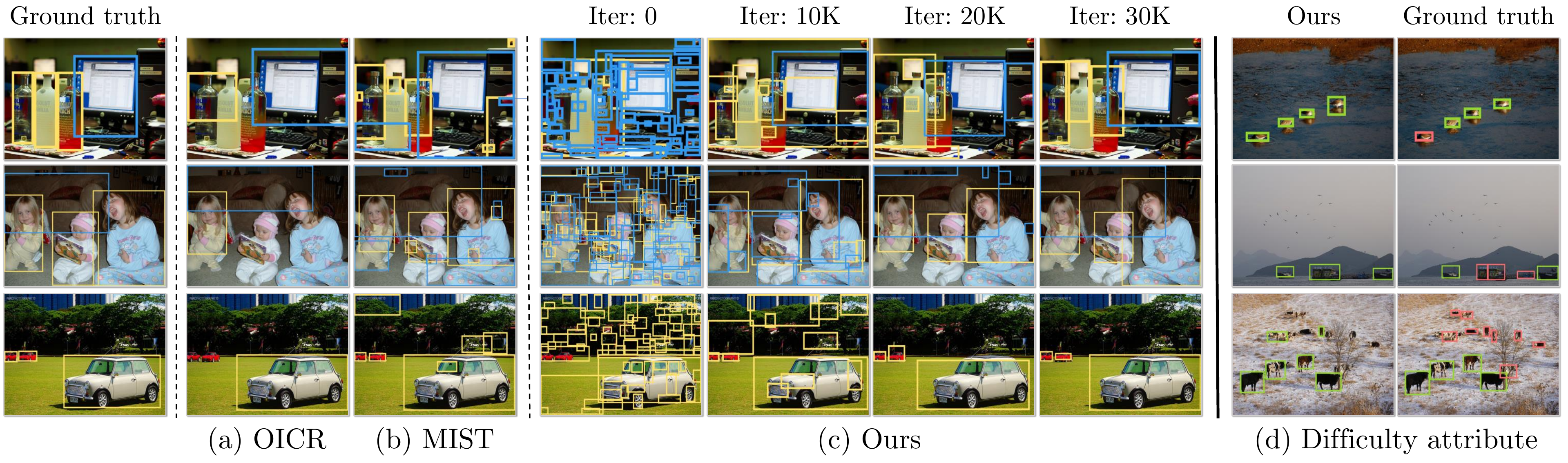}
\caption{
Comparison of pseudo groundtruths generated by (a) OICR~\cite{Tang_2017_CVPR}, (b) MIST~\cite{ren2020instance} and (c) Ours which shows pseudo groundtruths at different training steps.
``Difficult'' objects are often captured by Ours as shown in (d).}
\label{fig:figure6_pgt}
\end{figure}
\noindent\textbf{Selection of pseudo groundtruths.}
We visualize the pseudo groundtruths captured by OICR~\cite{Tang_2017_CVPR}, MIST~\cite{ren2020instance} and Ours in \cref{fig:figure6_pgt}.
OICR~\cite{Tang_2017_CVPR} selects only the top-scoring proposal per category ignoring all the other instances as shown in (a).
Although multiple objects are captured by MIST~\cite{ren2020instance} in (b), it also selects many false positives \eg object-like background.
Ours also captures many false positives in early stages of training (Iter: $0$ -- $10$K) but later in training, it mostly selects true positives (Iter: $20$K -- $30$K).
Our method can even detect some objects categorized as ``difficult'' (red boxes in (e)), which are not considered for detection performance as they are too hard even for humans to detect.

\section{Conclusion}
We propose a novel multiple instance labeling method to replace the conventional argmax-based pseudo groundtruth labeling method for weakly supervised object detection (WSOD).
To this end, we introduce a contrastive loss for the WSOD setting that learns consistent embedding features for proposals in the same class, and discriminative features for ones in different classes.
With these features, it is possible to mine a large number of reliable pseudo groundtruths, which provide richer supervision for WSOD tasks.
As a result, we achieve the new state-of-the-art results on both PASCAL VOC and MS-COCO benchmarks.

\smallskip
\noindent\textbf{Acknowledgements}.
This work was supported by Institute of Information \& communications Technology Planning \& Evaluation (IITP) grants funded by the Korea government (MSIT) (No.2017-0-00897, Development of Object Detection and Recognition for Intelligent Vehicles) and (No.B0101-15-0266, Development of High Performance Visual BigData Discovery Platform for Large-Scale Realtime Data Analysis),
as well as by support provided by the Natural Sciences and Engineering Research Council of Canada
and the Canada CIFAR AI Chairs program.
Junhyug Noh was supported by LLNL under Contract DE-AC52-07NA27344.

\clearpage
%
%
\bibliographystyle{splncs04}
\bibliography{eccv22_wsod}

\def\thesection{\Alph{section}}

\title{
[Appendix] \\
Object Discovery via Contrastive Learning for Weakly Supervised Object Detection}
\titlerunning{Object Discovery via Contrastive Learning for WSOD}
\author{Jinhwan Seo\inst{1}
\and
Wonho Bae\inst{2} 
\and
Danica J.\ Sutherland\inst{2,3} 
\and\\
Junhyug Noh\inst{4}\printfnsymbol{1}
\and
Daijin Kim\inst{1}\printfnsymbol{1}
}
\authorrunning{J. Seo et al.}
\institute{Pohang University of Science and Technology\and
University of British Columbia\and
Alberta Machine Intelligence Institute\and
Lawrence Livermore National Laboratory
\\
\email{tohoaa@gmail.com, whbae@cs.ubc.ca, dsuth@cs.ubc.ca, \\noh1@llnl.gov, dkim@postech.ac.kr}\\
\url{https://github.com/jinhseo/OD-WSCL}
}
\maketitle

In this appendix, we provide further results, both quantitative and qualitative, in the following order.
\begin{itemize}
    \item \cref{sec:performance_voc} reports  per-class Average Precision and Correct Localization results on PASCAL VOC datasets.

    \item \cref{sec:performance_proposal_generations} compares different proposal generation methods.
    
    \item \cref{sec:performance_cls_sim} demonstrates the robustness of proposed method using similarity threshold guided by WSCL. 
    
    \item \cref{sec:pseudo_code} provides overall pipeline of Object Discovery.
    
    \item \cref{sec:more_qual} provides additional qualitative results on PASCAL VOC and MS-COCO datasets.
    
\end{itemize}

\setcounter{equation}{6}
\setcounter{figure}{6}
\setcounter{table}{3}

\begin{figure*}[b!]
\centering
\includegraphics[width=1.0\textwidth]{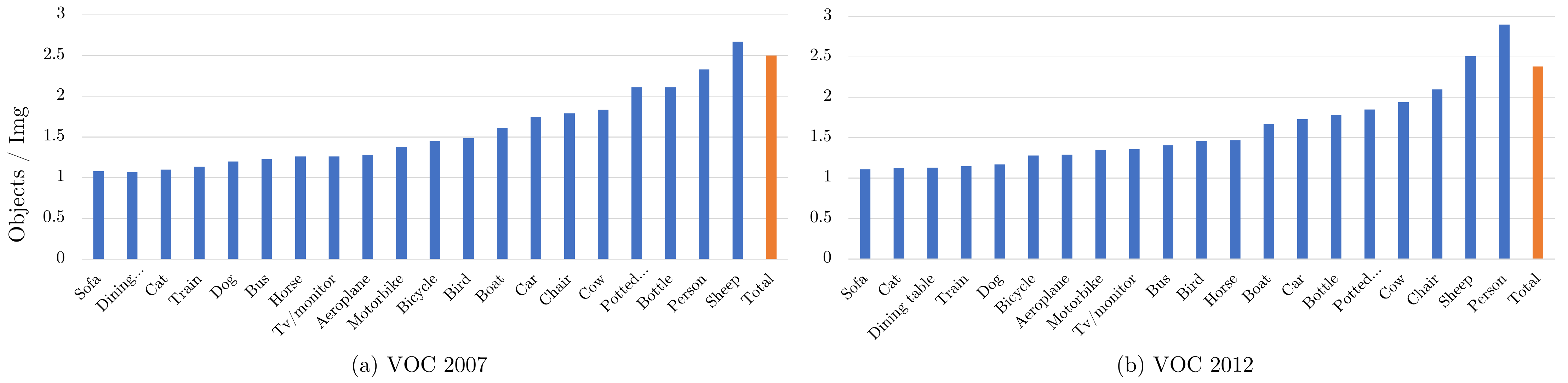}
\caption{Analysis of a number of objects per image on PASCAL VOC datasets.}
\label{fig:no_obj}
\vspace*{-10mm}
\end{figure*}

\section{Detailed Performance on PASCAL VOC}
\label{sec:performance_voc}
In \cref{tbl:class_ap_voc07,tbl:class_ap_voc12}, we provide additional performance of per-class average precision (AP) using Selective Search (SS)~\cite{uijlings2013selective} with VGG16 on VOC07 and VOC12~\cite{Everingham15}. 
Our method achieves the second-highest performance on VOC07 and the highest performance on VOC12.
The proposed method successfully addresses the issue of missing objects with high performance for the classes with a large number of objects per image, such as \textit{cow}, \textit{person} and \textit{sheep} in \cref{fig:no_obj}.

In \cref{tbl:class_corloc_voc07,tbl:class_corloc_voc12}, we report the results of per-class Correct Localization (CorLoc) scores using SS with VGG16 on VOC07 and VOC12. 
CorLoc is an additional evaluation metric commonly reported in WSOD literature to measure localization accuracy, equivalent to precision ($=\frac{\text{true positives}}{\text{true positives} + \text{false positives}}$).
More specifically, it measures the percentage of correct localization predictions where a prediction is treated as ``correct'' if the IoU between the prediction and corresponding ground truth is greater than or equal to $0.5$.
Our method achieves the third-best result in CorLoc on both VOC07 and VOC12.
Our slightly worse performance in CorLoc than in mAP is because, as a multiple instance labeling method, our approach captures more proposals than argmax-based methods: this significantly increases recall, but may slightly decrease precision (the only thing measured by CorLoc).

\begin{table*}[ht!]
\vspace*{-5mm}
\captionsetup{font=footnotesize}
\caption{Per-class AP results on VOC07}
\centering
\setlength{\tabcolsep}{3.5pt}
{\resizebox{\linewidth}{!}{
\begin{tabular}{c|cccccccccccccccccccc|c}
 \hline
 Method &Aero &Bike &Bird &Boat &Bottle &Bus &Car &Cat &Chair &Cow &Table &Dog &Horse &Motor &Person &Plant &Sheep &Sofa &Train &TV &mAP\\
 \hline\hline
 WSDDN\cite{Bilen_2016_CVPR}
 &39.4 &50.1 &31.5 &16.3 &12.6 &64.5 &42.8 &42.6 &10.1 &35.7 &24.9 &38.2 &34.4 &56.6 &9.4 &14.7 &30.2 &40.7 &54.7 &46.9 &34.8\\
 OICR\cite{Tang_2017_CVPR}
 &58.0 &62.4 &31.1 &19.4 &13.0 &65.1 &62.2 &28.4 &24.8 &44.7 &30.6 &25.3 &37.8 &65.5 &15.7 &24.1 &41.7 &46.9 &64.3 &62.6 &41.2\\
 C-WSL\cite{Gao_2018_ECCV}
 &62.9 &64.8 &39.8 &28.1 &16.4 &69.5 &68.2 &47.0 &27.9 &55.8 &43.7 &31.2 &43.8 &65.0 &10.9 &26.1 &52.7 &55.3 &60.2 &66.6 &46.8\\
 WSRPN\cite{Tang_2018_ECCV}
 &60.3 &66.2 &45.0 &19.6 &26.6 &68.1 &68.4 &49.4 &8.0 &56.9 &\textbf{55.0} &33.6 &62.5 &68.2 &20.6 &29.0 &49.0 &54.1 &58.8 &58.4 &47.9\\
 C-MIL\cite{Wan_2019_CVPR}
 &62.5 &58.4 &49.5 &32.1 &19.8 &70.5 &66.1 &63.4 &20.0 &60.5 &52.9 &53.5 &57.4 &68.9 &8.4 &24.6 &51.8 &58.7 &66.7 &63.5 &50.5\\
 C-MIDN\cite{gao2019c}
 &53.3 &71.5 &49.8 &26.1 &20.3 &70.3 &69.9 &68.3 &28.7 &65.3 &45.1 &\textbf{64.6} &58.0 &71.2 &20.0 &27.5 &54.9 &54.9 &69.4 &63.5 &52.6\\
 WSOD2\cite{zeng2019wsod2}
 &65.1 &64.8 &57.2 &39.2 &24.3 &69.8 &66.2 &61.0 &29.8 &64.6 &42.5 &60.1 &\textbf{71.2} &70.7 &21.9 &28.1 &58.6 &59.7 &52.2 &64.8 &53.6\\
 OIM\cite{sun2020mining}
 &55.6 &67.0 &45.8 &27.9 &21.1 &69.0 &68.3 &70.5 &21.3 &60.2 &40.3 &54.5 &56.5 &70.1 &12.5 &25.0 &52.9 &55.2 &65.0 &63.7 &50.1\\
 SLV\cite{chen2020slv}
 &65.6 &71.4 &49.0 &\textbf{37.1} &24.6 &69.6 &70.3 &70.6 &30.8 &63.1 &36.0 &61.4 &65.3 &68.4 &12.4 &\textbf{29.9} &52.4 &\textbf{60.0} &67.6 &64.5 &53.5\\
 MIST\cite{ren2020instance}
 &\textbf{68.8} &77.7 &57.0 &27.7 &\textbf{28.9} &69.1 &74.5 &67.0 &\textbf{32.1} &\textbf{73.2} &48.1 &45.2 &54.4 &73.7 &\textbf{35.0} &29.3 &\textbf{64.1} &53.8 &65.3 &65.2 &54.9\\
 CASD\cite{huang2020comprehensive}
 &- &- &- &- &- &- &- &- &- &- &- &- &- &- &- &- &- &- &- &- &\textbf{56.8}\\
 \hline
 Ours    &65.8 &\textbf{79.5} &\textbf{58.1} &23.7 &28.6 &\textbf{71.2} &\textbf{75.0} &\textbf{71.7} &31.7 &69.8 &45.2 &55.7 &57.2 &\textbf{75.7} &29.6 &24.3 &61.0 &55.3 &\textbf{71.7} &\textbf{72.0} &56.1\\
 \Xhline{2\arrayrulewidth}
\end{tabular}
}}
\label{tbl:class_ap_voc07}
\vspace*{-15mm}
\end{table*}

\begin{table*}[h!]
\captionsetup{font=footnotesize}
\caption{Per-class AP results on VOC12}
\centering
\setlength{\tabcolsep}{3.5pt}
{\resizebox{\linewidth}{!}{
\begin{tabular}{c|cccccccccccccccccccc|c}
 \hline
 Method &Aero &Bike &Bird &Boat &Bottle &Bus &Car &Cat &Chair &Cow &Table &Dog &Horse &Motor &Person &Plant &Sheep &Sofa &Train &TV &mAP\\
 \hline\hline
 OICR\cite{Tang_2017_CVPR}
 &67.7 &61.2 &41.5 &25.6 &22.2 &54.6 &49.7 &25.4 &19.9 &47.0 &18.1 &26.0 &38.9 &67.7 &2.0 &22.6 &41.1 &34.3 &37.9 &55.3 &37.9\\
 C-WSL\cite{Gao_2018_ECCV}
 &74.0 &67.3 &45.6 &29.2 &26.8 &62.5 &54.8 &21.5 &22.6 &50.6 &24.7 &25.6 &57.4 &71.0 &2.4 &22.8 &44.5 &44.2 &45.2 &\textbf{66.9} &43.0\\
 WSRPN\cite{Tang_2018_ECCV}
 &- &- &- &- &- &- &- &- &- &- &- &- &- &- &- &- &- &- &- &- &43.4\\
 C-MIL\cite{Wan_2019_CVPR}
 &- &- &- &- &- &- &- &- &- &- &- &- &- &- &- &- &- &- &- &- &46.7\\
 C-MIDN\cite{gao2019c}
 &72.9 &68.9 &53.9 &25.3 &29.7 &60.9 &56.0 &\textbf{78.3} &23.0 &57.8 &\textbf{25.7} &\textbf{73.0} &\textbf{63.5} &73.7 &13.1 &28.7 &51.5 &35.0 &56.1 &57.5 &50.2\\
 WSOD2\cite{zeng2019wsod2}
 &- &- &- &- &- &- &- &- &- &- &- &- &- &- &- &- &- &- &- &- &47.2\\
 OIM\cite{sun2020mining}
 &- &- &- &- &- &- &- &- &- &- &- &- &- &- &- &- &- &- &- &- &45.3\\
 SLV\cite{chen2020slv}
 &- &- &- &- &- &- &- &- &- &- &- &- &- &- &- &- &- &- &- &- &49.2\\
 MIST\cite{ren2020instance}
 &\textbf{78.3} &73.9 &56.5 &30.4 &37.4 &64.2 &59.3 &60.3 &\textbf{26.6} &66.8 &25.0 &55.0 &61.8 &79.3 &14.5 &30.3 &61.5 &40.7 &56.4 &63.5 &52.1\\
 CASD\cite{huang2020comprehensive}
 &- &- &- &- &- &- &- &- &- &- &- &- &- &- &- &- &- &- &- &- &53.6\\
 \hline
 Ours
 &73.8 &\textbf{74.7} &\textbf{61.3} &\textbf{32.9} &\textbf{40.0} &\textbf{64.6} &\textbf{59.8} &68.1 &26.3 &\textbf{67.5} &23.0 &67.1 &62.8 &\textbf{80.6} &\textbf{17.3} &\textbf{34.1} &\textbf{63.4} &\textbf{44.4} &\textbf{66.2} &64.9 &\textbf{54.6}\\
 \Xhline{2\arrayrulewidth}
\end{tabular}
}}
\label{tbl:class_ap_voc12}
\vspace*{-15mm}
\end{table*}
\begin{table*}[ht!]
\captionsetup{font=footnotesize}
\caption{Per-class CorLoc results on VOC07}
\centering
\setlength{\tabcolsep}{3.5pt}
{\resizebox{\linewidth}{!}{
\begin{tabular}{c|cccccccccccccccccccc|c}
 \hline
 Method &Aero &Bike &Bird &Boat &Bottle &Bus &Car &Cat &Chair &Cow &Table &Dog &Horse &Motor &Person &Plant &Sheep &Sofa &Train &TV &CorLoc\\
 \hline\hline
 WSDDN\cite{Bilen_2016_CVPR}
 &65.1 &58.8 &58.5 &33.1 &39.8 &68.3 &60.2 &59.6 &34.8 &64.5 &30.5 &43.0 &56.8 &82.4 &25.5 &41.6 &61.5 &55.9 &65.9 &63.7 &53.5\\
 OICR\cite{Tang_2017_CVPR}
 &81.7 &80.4 &48.7 &49.5 &32.8 &81.7 &85.4 &40.1 &40.6 &79.5 &35.7 &33.7 &60.5 &88.8 &21.8 &57.9 &76.3 &59.9 &75.3 &81.4 &60.6\\
 C-WSL\cite{Gao_2018_ECCV}
 &85.8 &81.2 &64.9 &50.5 &32.1 &84.3 &85.9 &54.7 &43.4 &80.1 &42.2 &42.6 &60.5 &90.4 &13.7 &57.5 &\textbf{82.5} &61.8 &74.1 &82.4 &63.5\\
 WSRPN\cite{Tang_2018_ECCV}
 &77.5 &81.2 &55.3 &19.7 &44.3 &80.2 &86.6 &69.5 &10.1 &87.7 &\textbf{68.4} &52.1 &84.4 &91.6 &\textbf{57.4} &\textbf{63.4} &77.3 &58.1 &57.0 &53.8 &63.8\\
 C-MIL\cite{Wan_2019_CVPR}
 &- &- &- &- &- &- &- &- &- &- &- &- &- &- &- &- &- &- &- &- &65.0\\
 C-MIDN\cite{gao2019c}
 &- &- &- &- &- &- &- &- &- &- &- &- &- &- &- &- &- &- &- &- &68.7\\
 WSOD2\cite{zeng2019wsod2}
 &87.1 &80.0 &74.8 &\textbf{60.1} &36.6 &79.2 &83.8 &70.6 &43.5 &\textbf{88.4} &46.0 &74.7 &87.4 &90.8 &44.2 &52.4 &81.4 &61.8 &67.7 &79.9 &69.5\\
 OIM\cite{sun2020mining}
 &- &- &- &- &- &- &- &- &- &- &- &- &- &- &- &- &- &- &- &- &67.2\\
 SLV\cite{chen2020slv}
 &84.6 &84.3 &73.3 &58.5 &\textbf{49.2} &80.2 &87.0 &79.4 &46.8 &83.6 &41.8 &\textbf{79.3} &\textbf{88.8} &90.4 &19.5 &59.7 &79.4 &\textbf{67.7} &\textbf{82.9} &\textbf{83.2} &\textbf{71.0}\\
 MIST\cite{ren2020instance}
 &\textbf{87.5} &82.4 &\textbf{76.0} &58.0 &44.7 &82.2 &\textbf{87.5} &71.2 &\textbf{49.1} &81.5 &51.7 &53.3 &71.4 &\textbf{92.8} &38.2 &52.8 &79.4 &61.0 &78.3 &76.0 &68.8\\
 CASD\cite{huang2020comprehensive}
 &- &- &- &- &- &- &- &- &- &- &- &- &- &- &- &- &- &- &- &- &70.4\\
 \hline
 Ours
 &86.3 &\textbf{87.8} &74.5 &47.3 &43.9 &\textbf{85.8} &84.6 &\textbf{78.2} &\textbf{49.1} &83.6 &49.4 &61.6 &74.5 &92.4 &42.2 &46.9 &80.4 &62.1 &\textbf{82.9} &82.8 &69.8\\
 \Xhline{2\arrayrulewidth}
\end{tabular}
}}
\label{tbl:class_corloc_voc07}
\vspace*{-15mm}
\end{table*}

\begin{table*}[h!]
\captionsetup{font=footnotesize}
\caption{Per-class CorLoc results on VOC12}
\centering
\setlength{\tabcolsep}{3.5pt}
{\resizebox{\linewidth}{!}{
\begin{tabular}{c|cccccccccccccccccccc|c}
 \hline
 Method &Aero &Bike &Bird &Boat &Bottle &Bus &Car &Cat &Chair &Cow &Table &Dog &Horse &Motor &Person &Plant &Sheep &Sofa &Train &TV &CorLoc\\
 \hline\hline
 OICR\cite{Tang_2017_CVPR}
 &- &- &- &- &- &- &- &- &- &- &- &- &- &- &- &- &- &- &- &-  &62.1\\
 C-WSL\cite{Gao_2018_ECCV}
 &90.9 &81.1 &64.9 &57.6 &50.6 &84.9 &78.1 &29.8 &49.7 &83.9 &50.9 &42.6 &78.6 &87.6 &10.4 &58.1 &85.4 &\textbf{61.0} &64.7 &\textbf{86.6} &64.9\\
 WSRPN\cite{Tang_2018_ECCV}
 &85.5 &60.8 &62.5 &36.6 &53.8 &82.1 &80.1 &48.2 &14.9 &87.7 &\textbf{68.5} &60.7 &\textbf{85.7} &89.2 &\textbf{62.9} &\textbf{62.1} &87.1 &54.0 &45.1 &70.6 &64.9\\
 C-MIL\cite{Wan_2019_CVPR}
 &- &- &- &- &- &- &- &- &- &- &- &- &- &- &- &- &- &- &- &- &67.4\\
 C-MIDN\cite{gao2019c}
 &- &- &- &- &- &- &- &- &- &- &- &- &- &- &- &- &- &- &- &- &71.2\\
 WSOD2\cite{zeng2019wsod2}
 &- &- &- &- &- &- &- &- &- &- &- &- &- &- &- &- &- &- &- &- &71.9\\
 OIM\cite{sun2020mining}
 &- &- &- &- &- &- &- &- &- &- &- &- &- &- &- &- &- &- &- &- &67.1\\
 SLV\cite{chen2020slv}
 &- &- &- &- &- &- &- &- &- &- &- &- &- &- &- &- &- &- &- &- &69.2\\
 MIST\cite{ren2020instance}
 &\textbf{91.7} &85.6 &71.7 &56.6 &55.6 &88.6 &\textbf{77.3} &\textbf{63.4} &\textbf{53.6} &\textbf{90.0} &51.6 &62.6 &79.3 &\textbf{94.2} &32.7 &58.8 &90.5 &57.7 &\textbf{70.9} &85.7 &70.9\\
 CASD\cite{huang2020comprehensive}
 &- &- &- &- &- &- &- &- &- &- &- &- &- &- &- &- &- &- &- &- &\textbf{72.3}\\
 \hline
 Ours
 &88.2 &\textbf{88.3} &\textbf{75.0} &\textbf{59.7} &\textbf{58.9} &\textbf{89.3} &73.2 &57.8 &53.4 &88.0 &48.7 &\textbf{67.5} &78.3 &94.0 &34.8 &61.6 &\textbf{91.7} &59.4 &\textbf{70.9} &84.4 &71.2\\
 \Xhline{2\arrayrulewidth}
\end{tabular}
}}
\label{tbl:class_corloc_voc12}
\vspace*{-15mm}
\end{table*}

\clearpage
\section{Comparison of Proposal Generation Method}
\label{sec:performance_proposal_generations}
Current WSOD models rely on pre-computed proposal methods such as Selective Search (SS)~\cite{uijlings2013selective} and Edge Boxes (EB)~\cite{zitnick2014edge}.
Although the choice of proposal generation methods has a significant impact on localization performance,
most previous studies still exploit SS for PASCAL VOC and MCG for MS-COCO datasets.
To better understand the effect of using different proposal methods, we compare our algorithm's performance to that of several state-of-the-art algorithms with different proposal methods (SS~\cite{uijlings2013selective}, MCG~\cite{arbelaez2014multiscale}, and COB~\cite{maninis2016convolutional}) on VOC07 (\cref{tbl:proposals_voc07}) and MS-COCO (\cref{tbl:proposal_coco}) datasets.
Note that COB generally captures the groundtruths the best among the three proposal generation methods whereas SS performs the worst.

In general, the better the proposals are, the higher the performance of detection is regardless of model.
In \cref{tbl:proposals_voc07}, Ours performs the best with COB and then with MCG (COB: $61.8\%$, MCG: $58.7\%$, and SS: $56.1\%$), which is the same for CASD and MIST.
Similarly, COB outperforms MCG with a large margin as observed on MS-COCO datasets as shown in \cref{tbl:proposal_coco}.
We chose to report only the performance of SS and MCG in the main paper because additional boundary information is required to train COB, which violates the definition of image-level supervision.
Based on this experiment, we believe MCG should be the default proposal generation method for both PASCAL VOC and MS-COCO datasets unlike the previous convention in WSOD.

\begin{table*}[ht!]
\vspace{-3mm}
\captionsetup{font=footnotesize}
\caption{Per-class AP results with different proposal generation methods on VOC07}
\centering
\setlength{\tabcolsep}{3.5pt}
{\resizebox{\linewidth}{!}{
\begin{tabular}{c|c|cccccccccccccccccccc|c}
 \hline
 Method &Proposal&Aero &Bike &Bird &Boat &Bottle &Bus &Car &Cat &Chair &Cow &Table &Dog &Horse &Motor &Person &Plant &Sheep &Sofa &Train &TV &mAP\\
 \hline\hline
 MIST\cite{ren2020instance}
 &SS &68.8 &77.7 &57.0 &27.7 &28.9 &69.1 &74.5 &67.0 &32.1 &73.2 &48.1 &45.2 &54.4 &73.7 &35.0 &29.3 &64.1 &53.8 &65.3 &65.2 &54.9\\
 CASD\cite{huang2020comprehensive}
 &SS &- &- &- &- &- &- &- &- &- &- &- &- &- &- &- &- &- &- &- &- &56.8\\
 Ours    &SS &65.8 &79.5 &58.1 &23.7 &28.6 &71.2 &75.0 &71.7 &31.7 &69.8 &45.2 &55.7 &57.2 &75.7 &29.6 &24.3 &61.0 &55.3 &71.7 &72.0 &56.1\\
 \hline
 MIST\cite{ren2020instance}
 &MCG &65.7 &78.9 &55.5 &25.1 &31.3 &74.5 &76.8 &67.5 &16.1 &68.7 &50.3 &36.0 &73.4 &76.7 &31.7 &30.7 &61.6 &64.5 &74.9 &70.0 &56.5\\
 CASD\cite{huang2020comprehensive}
 &MCG &65.1 &70.5 &55.6 &42.8 &31.3 &72.4 &71.7 &75.5 &16.0 &64.1 &\textbf{60.2} &68.4 &71.5 &70.7 &39.6 &27.5 &58.3 &53.9 &63.6 &69.2 &57.4\\
 Ours    &MCG &\textbf{69.2} &\textbf{81.5} &56.4 &28.5 &30.5 &77.6 &79.1 &71.6 &13.0 &70.8 &48.8 &56.9 &74.9 &\textbf{78.4} &34.9 &27.6 &61.4 &\textbf{65.4} &74.4 &\textbf{73.4} &58.7\\
 \hline
 MIST\cite{ren2020instance}
 &COB &65.1 &74.8 &57.5 &34.0 &45.0 &77.8 &\textbf{80.6} &56.1 &20.5 &\textbf{71.2} &50.0 &51.9 &58.0 &78.2 &27.2 &\textbf{32.6} &\textbf{62.2} &63.4 &72.9 &69.8 &57.4\\
 CASD\cite{huang2020comprehensive}
 &COB &69.1 &71.1 &\textbf{63.2} &\textbf{48.5} &40.0 &76.4 &74.2 &\textbf{77.1} &17.6 &67.4 &59.9 &\textbf{76.1} &74.4 &70.4 &20.8 &30.2 &59.4 &58.3 &67.2 &68.1 &59.4\\
 Ours    &COB &68.6 &78.4 &62.2 &36.6 &\textbf{49.8} &\textbf{79.2} &80.9 &77.0 &\textbf{29.4} &71.0 &38.1 &62.7 &\textbf{80.6} &78.0 &\textbf{40.8} &31.6 &61.7 &62.8 &\textbf{75.7} &69.8 &\textbf{61.8}\\
 \Xhline{2\arrayrulewidth}
\end{tabular}
}}
\label{tbl:proposals_voc07}
\vspace{-7mm}
\end{table*}

\begin{table*}[ht!]
\captionsetup{font=footnotesize}
\caption{Performance with different proposal generation methods on MS-COCO}
\centering
\setlength{\tabcolsep}{4.0pt}
\resizebox{\textwidth}{!}{
\begin{tabular}{c|c|c|c|c|cc|ccc|ccc|ccc}
 \hline
 Dataset &Backbone &Method &Proposal &$AP$ &$AP^{50}$ &$AP^{75}$ &$AP^s$ &$AP^m$ &$AP^l$ &$AR^1$ &$AR^{10}$ &$AR^{100}$ &$AR^s$ &$AR^m$ &$AR^l$\\
 \hline\hline
 \multirow{11}{*}{COCO14} & \multirow{4}{*}{VGG16}  
              &MIST~\cite{ren2020instance}   &MCG &11.4 &24.3 &9.4 &3.6 &12.2 &17.6 &13.5 &22.6 &23.9 &8.5 &25.4 &38.3 \\
 &               &CASD~\cite{huang2020comprehensive}   &MCG &12.8 &26.4 &-&-&-&-&-&-&-&-&-&-\\
 &               &Ours &MCG  &13.7 &27.7 &11.9 &4.4 &14.5 &21.2 &14.7 &24.8 &26.9 &8.8 &27.8 &44.0  \\
 &                        &Ours   &COB
 &\textbf{15.1} &\textbf{29.3} &\textbf{13.8} &\textbf{4.5} &\textbf{15.9} &\textbf{23.4} &\textbf{16.0} &\textbf{26.5}
 &\textbf{28.2} &\textbf{8.9} &\textbf{29.5} &\textbf{46.5}\\
\cline{2-16}
 & \multirow{4}{*}{ResNet50} &MIST~\cite{ren2020instance} &MCG &12.6 &26.1 &10.8 &3.7 &13.3 &19.9 &14.8 &23.7 &24.7 &8.4 &25.1 &41.8 \\
 &       &CASD~\cite{huang2020comprehensive}  &MCG &13.9      &27.8     &-           &-   &-&-&-&-&-&-&-&- \\
 &       &Ours &MCG
 &13.9 &29.1 &11.8 &\textbf{4.9} &16.8 &22.3 &15.5 &26.1 &28.0 &9.0 &31.8 &46.6 \\
 &       &Ours &COB
 &\textbf{15.4} &\textbf{30.4} &\textbf{14.0} &4.8 &\textbf{18.0} &\textbf{24.6} &\textbf{16.9} &\textbf{29.2} &\textbf{31.4} &\textbf{9.4} &\textbf{35.1} &\textbf{53.1} \\
 \cline{2-16}
 &\multirow{3}{*}{ResNet101}      &MIST~\cite{ren2020instance} &MCG
 &13.0      &26.1     &10.8         &3.7   &13.3 &19.9 &14.8 &23.7 &24.7 &8.4 &25.1 &41.8 \\
 &       &Ours     &MCG
 &14.4 &29.0 &12.4 &4.8 &17.3 &23.8 &15.8 &27.0 &30.0 &9.2 &33.6 &51.0\\
 &       &Ours     &COB
 &\textbf{16.2} &\textbf{31.6} &\textbf{14.8} &\textbf{5.0} &\textbf{18.7} &\textbf{26.4} &\textbf{17.5} &\textbf{29.6}
 &\textbf{31.9} &\textbf{10.0} &\textbf{35.4} &\textbf{53.5}\\
 \hline
 \multirow{7}{*}{COCO17}  &\multirow{3}{*}{VGG16}      &MIST~\cite{ren2020instance} &MCG
 &12.4      &25.8     &10.5        &3.9  &13.8 &19.9 &14.3 &23.3 &24.6 &9.7 &26.6 &39.6 \\
 &                              &Ours  &MCG
 &13.6 &27.4 &12.2 &4.9 &15.5 &21.6 &14.6 &24.8 &26.8 &9.2 &28.7 &43.8 \\
 &         &Ours &COB
 &\textbf{15.6} &\textbf{29.9} &\textbf{14.3} &\textbf{5.1} &\textbf{17.2} &\textbf{25.1} &\textbf{16.4} &\textbf{27.1}
 &\textbf{28.7} &\textbf{9.8} &\textbf{30.5} &\textbf{47.8} \\
\cline{2-16}
 &\multirow{2}{*}{ResNet50}       &Ours &MCG &13.8 &27.8 &12.1 &\textbf{5.7} &17.7 &23.8 &15.1 &26.6 &29.7 &10.1 &33.7 &50.7\\
 &               &Ours &COB 
 &\textbf{16.0} &\textbf{30.5} &\textbf{14.9} &5.4 &\textbf{19.0} &\textbf{27.2} &\textbf{17.0} &\textbf{29.1}
 &\textbf{31.4} &\textbf{10.4} &\textbf{35.2} &\textbf{53.3}\\
\cline{2-16}
 &\multirow{2}{*}{ResNet101}      &Ours &MCG &14.4 &28.7 &12.6 &5.4 &17.9 &25.5 &15.4 &26.8 &29.6 &10.0 &33.3 &50.6\\
 &               &Ours &COB &\textbf{16.5} &\textbf{31.6} &\textbf{15.2} &\textbf{5.7} &\textbf{19.6} &\textbf{28.2} &\textbf{17.4} &\textbf{29.7} &\textbf{31.9} &\textbf{11.3} &\textbf{35.5} &\textbf{54.2}\\
 \Xhline{2\arrayrulewidth}
\end{tabular}}
\label{tbl:proposal_coco}
\end{table*}

\section{Different Criterion for Object Discovery}
\label{sec:performance_cls_sim}
In Section 4.3, we claimed that the similarity of two proposals in the embedding space can be large even though they are not similar in classification score.
To justify the necessity of using additional similarity scores for the object discovery module, \cref{tbl:sim_cls} compares object discovery based on classification score, with various threshold values, to similarity score.
Recall that the ``adaptive'' threshold we used for similarity score is determined by the average value of similarity between the argmax and its augmented samples: $\tau^{n,k}_c = \frac{1}{|\mathcal{S}_c|}\sum^{|\mathcal{S}_c|}_{i=1}{sim(z^{n}_{\bar{m}^{n,k}_c}, \mathcal{S}_{c,i})}$. 
For object discovery based on classification score, we not only try fixed thresholds but also adaptive threshold defined as $\tau^{n,k}_c = \frac{1}{|\mathcal{S'}_c|}\sum^{|\mathcal{S'}_c|}_{i=1}{\mathcal{S'}_{c,i}}$ where $\mathcal{S'}$ is the collection of classification scores that are calculated using the augmented features (same features for    $\mathcal{S}$).

In \cref{tbl:sim_cls}, the performance of the object discovery based on classification score is significantly worse than similarity score.
Moreover,
the best-performing threshold $\tau^{n,k}_c = 0.4$ ($57.4\%$) is dramatically better than a similar threshold value $\tau^{n,k}_c = 0.2$.
Thus, unlike similarity score (as shown in Section 5.3), performance is also very sensitive to the choice of threshold.
Note that we train the model with the same hyperparameters ($\tau_{nms} = 0.1$, $\lambda = 0.03$) for fair comparison.

\begin{table*}[h!]
    \vspace*{-8mm}
    \captionsetup{font=footnotesize}
    \caption{The results of different criteria for object discovery}
    \vspace{1mm}
    \centering
    \begin{tabular}[t]{c|c|D}
        \firsthline
        Criterion &Threshold ($\tau^{n,k}_c$)  &mAP\\ 
        \hline\hline
        \multirow{8}{*}{Classification Score} &0.2     &50.3\\
        &0.3     &56.2\\
        &0.4     &57.4 \\
        &0.5     &56.2 \\
        &0.6     &56.1 \\
        &0.7     &55.6 \\
        &0.8     &54.4 \\
        &Adaptive&53.2 \\
        \hline
        Similarity Score &Adaptive&\textbf{58.7}\\
        \Xhline{2\arrayrulewidth}
    \end{tabular}
\label{tbl:sim_cls}
\vspace*{-3mm}
\end{table*}

\begin{figure*}[b!]
\centering
\includegraphics[width=1.0\textwidth]{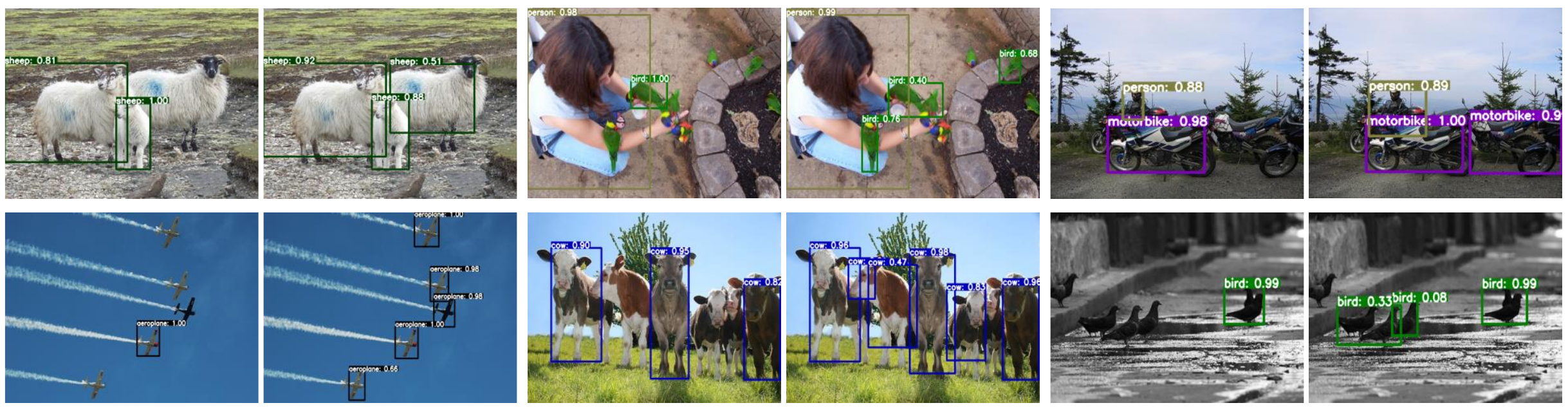}
\caption{Comparison of pseudo groundtruths generated by classification score \vs{ } similarity score. 
The left and right images of each pair correspond to pseudo groundtruths based on classification and similarity scores, respectively.
}
\label{fig:more_pgt}
\end{figure*}

\clearpage
\section{Pseudo Code: Sampling and Object Discovery}
Along with Fig.2 in the main paper, we provide the detailed procedure of sampling steps and object discovery. The main purpose of object discovery is to obtain more pseudo groundtruths in addition to the top-scoring proposals.
\label{sec:pseudo_code}
\begin{algorithm}[H]
\caption{Sampling steps and object discovery}
\textbf{Network}: RoI feature extractor $\eta(\cdot)$, similarity head $\varphi(\cdot)$\\
\textbf{Input}:  Proposal Scores: $x^{n,k}_{c,m}$, Embedding Vectors: $z^{n}_m$,\\
Proposals: $R^n$, Image Labels: $Y^n$, Proposal Labels: $Y^{n,k}_{c,m}$\\
\textbf{Output}: Updated $S_c$, $Y^{n,k}_{c,m}$
\begin{algorithmic}[1]
    \STATE $S_c \leftarrow{\emptyset}, y^{n,k}_{c,m} = 0, y^{n,k}_{(c+1),m} = 1$
    \FOR {n = 1 \textbf{to} N}
      \FOR {k = 0 \textbf{to} K-1} 
          \IF {$y^{n}_c$==1}
          \STATE $\bar{m}^{n,k}_c=\argmax_m {x^{n,(k-1)}_{c,m}}$
            \IF {$IoU(r_m, r_{\bar{m}^{n,k}_c}) > \tau_{IoU}, \forall{m}\in{M^n}$}
                \STATE $\mathcal{M}^{n,k}_c \leftarrow{m}$
            \ENDIF
          \ENDIF
      \ENDFOR
      \STATE $\mathcal{Z}^{n,c}_{IoU} = \{ \varphi(\eta{(f^n_m)})\hspace{0.1cm} | \hspace{0.1cm} m \in \bigcup_{k=0}^{K-1} \mathcal{M}^{n,k}_c \}$
      \STATE $D: D_{i,j}\sim U(0,1) \in \mathbb{R}^{H \times W}$
      \STATE $D_{drop} = \begin{cases}
                        0  & \text{if $D < \tau_{drop}$}\\
                        1  & \text{otherwise}
                        \end{cases}$
      \STATE $\mathcal{Z}^{n,c}_{mask} = \{ \varphi(\eta{(f^n_m \odot D_{drop})})\hspace{0.1cm} | \hspace{0.1cm} m \in \bigcup_{k=0}^{K-1} \mathcal{M}^{n,k}_c \}$
      \STATE $D_{noise}: D_{i,j} \sim N(0,1) \in \mathbb{R}^{H \times W}$
      \STATE $\mathcal{Z}^{n,c}_{noise} = \{ \varphi(\eta(f^n_m + f^n_m \odot D_{noise}))\hspace{0.1cm} | \hspace{0.1cm} m \in \bigcup_{k=0}^{K-1} \mathcal{M}^{n,k}_c \}$
    \ENDFOR
    \STATE $\mathcal{S}_c = \bigcup_{n=1}^N ( \mathcal{Z}^{n,c}_{IoU} \cup \mathcal{Z}^{n,c}_{mask} \cup \mathcal{Z}^{n,c}_{noise} )$
    \FOR {n = 1 \textbf{to} N}
      \FOR {k = 0 \textbf{to} K-1}
        \IF {$y^{n}_c$==1}
            \STATE $\bar{m}^{n,k}_c=\argmax_m {x^{n,(k-1)}_{c,m}}$
            \STATE $\tau^n_c = Avg(sim(z^n_{\bar{m}^{n,k}_c}, S_{c}))$
            \IF {$sim(z^n_{\bar{m}^{n,k}_c}, z^n_{m}) > \tau^n_c, \forall{m}\in{M^n}$}
                \STATE $S_c  \leftarrow{z^{n}_{m}}$
                \IF {$IoU(r_m, r_{\bar{m}^{n,k}_c}) > 0.5, \forall{m}\in{M^n}$}
                  \STATE $y^{n,k}_{c,m} = 1$
                \ENDIF
            \ENDIF
        \ENDIF
      \ENDFOR
    \ENDFOR
\end{algorithmic}
\label{alg:OD}
\end{algorithm}

\clearpage
\section{More Qualitative Results}  
\label{sec:more_qual}
\begin{figure*}[ht!]
\centering
\includegraphics[width=1.0\textwidth]{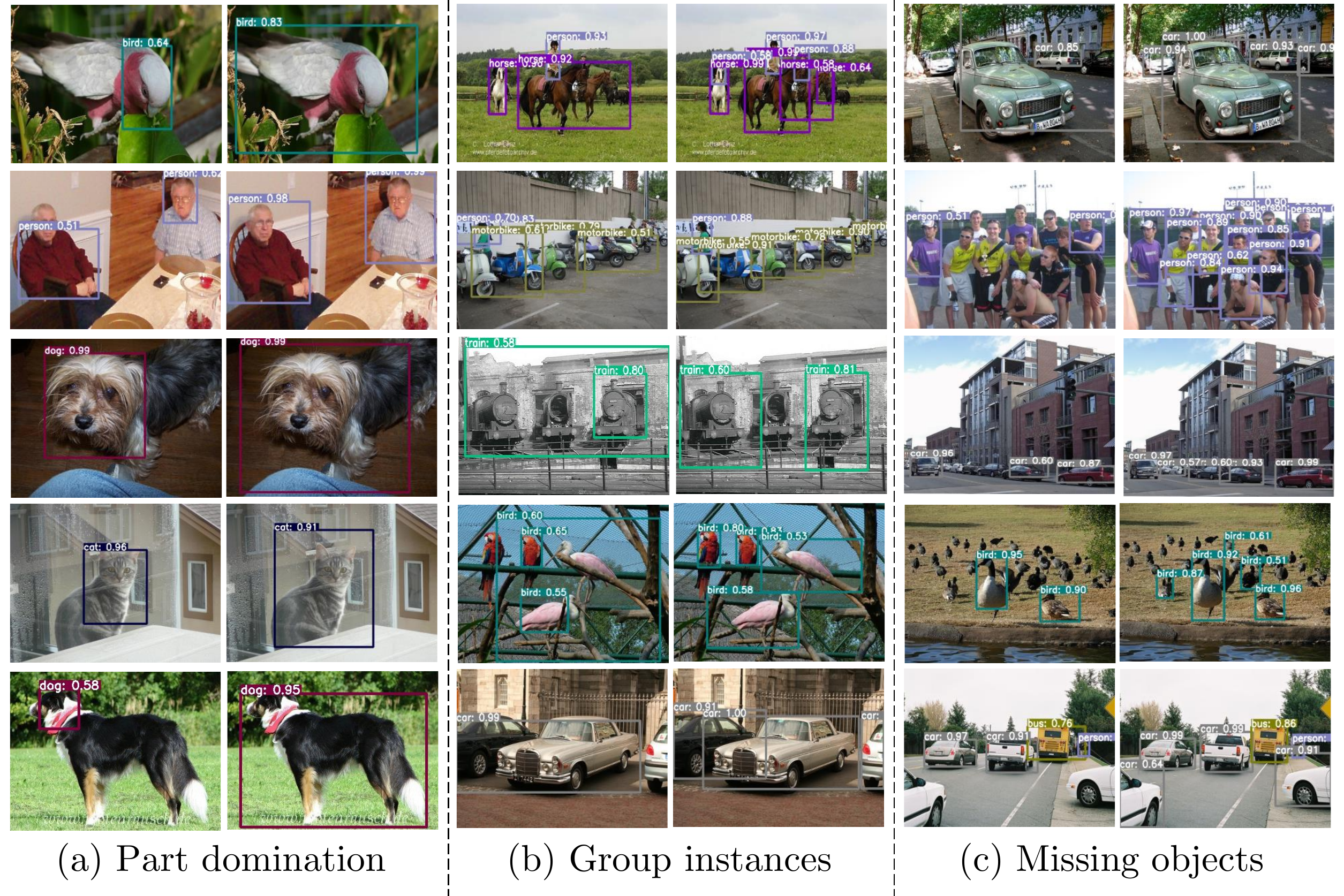}
\caption{More qualitative results for the three challenges of WSOD on VOC07.}
\label{fig:more_vis_voc}
\end{figure*}

In \cref{fig:more_vis_voc}, we provide more qualitative results for the three challenges of WSOD on VOC07. 
Columns on the left and right of each pair correspond to qualitative results from OICR~\cite{Tang_2017_CVPR} and our model, respectively.

\begin{figure*}[ht!]
\centering
\includegraphics[width=1.0\textwidth]{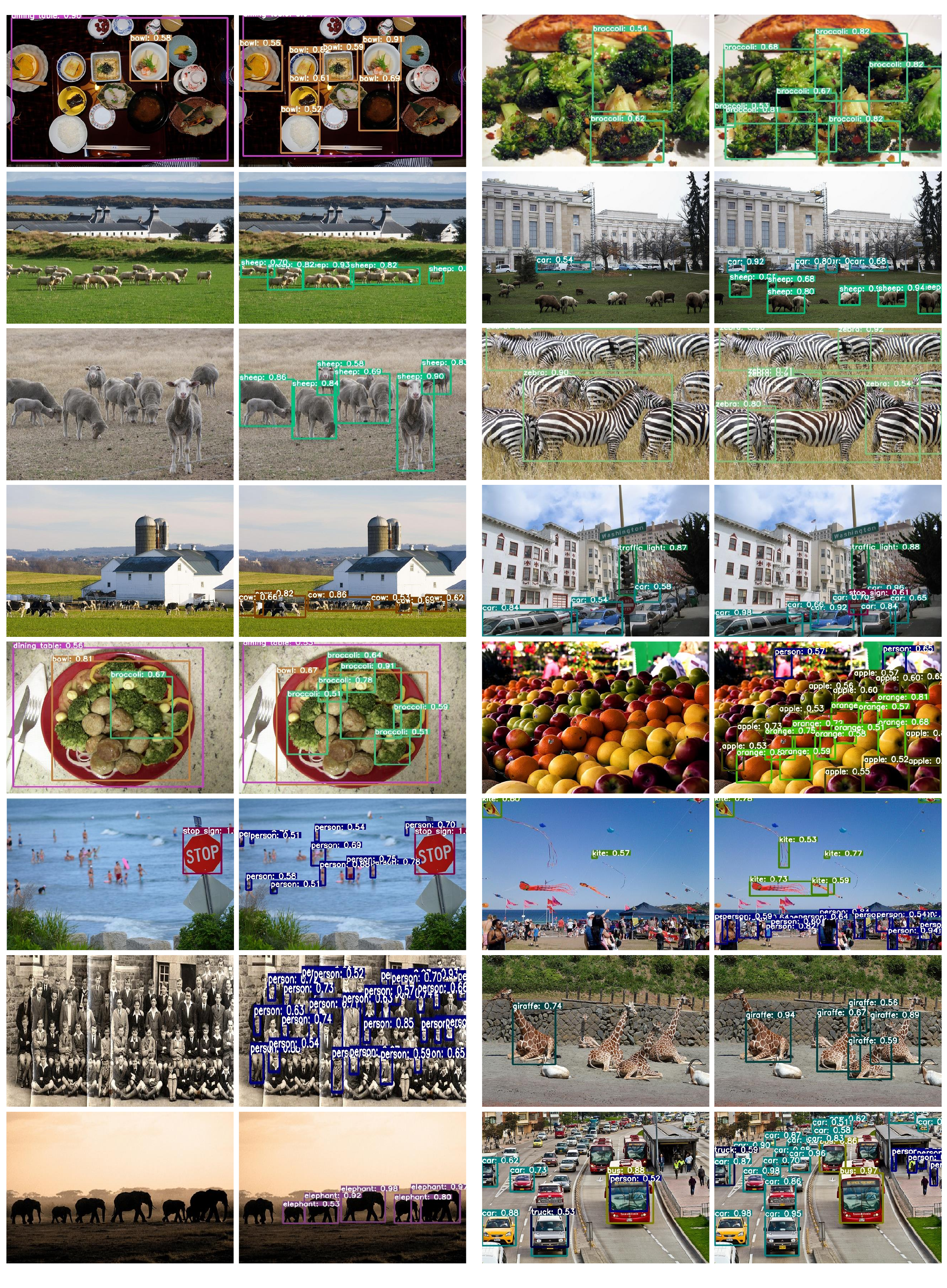}
\caption{Qualitative results on COCO14.}
\label{fig:more_vis_coco}
\end{figure*}

In \cref{fig:more_vis_coco}, we compare prediction results of OICR~\cite{Tang_2017_CVPR} on the left and Ours on the right.  
Our model shows much better results for COCO, which contains more instances per image. Although the issue of grouped instances is observed in some cases, our model correctly captures multiple objects and classifies them correctly, despite extremely complex backgrounds.

\begin{figure*}[ht!]
\centering
\includegraphics[width=1.0\textwidth]{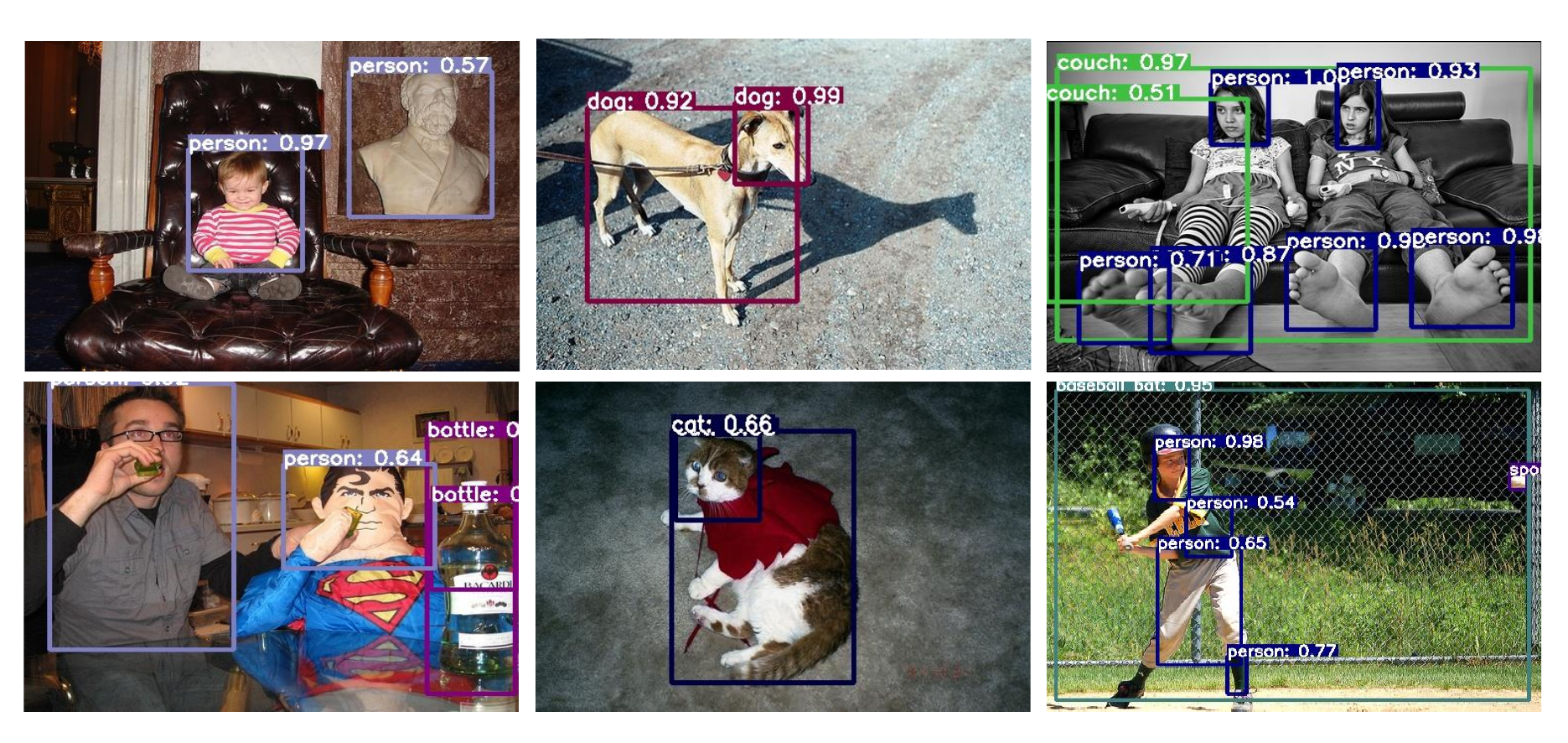}
\caption{Failure cases of the proposed method.
}
\label{fig:failure}
\end{figure*}

\cref{fig:failure} shows failure cases of the proposed method. Our model misclassfies background objects that looks like a target class, for example human-like statues or dolls. In addition, the predicted boxes are separated in some cases, even though the object its full extent is captured.

\clearpage



%
%

\end{document}